\pdfoutput=1
\documentclass[11pt]{article}

\usepackage[final]{acl}

\usepackage{times}
\usepackage{latexsym}

\usepackage[T1]{fontenc}
\usepackage{booktabs}%
\usepackage{multirow}%
\usepackage{array}%

\usepackage{siunitx}%
\usepackage{tabularx}%
\usepackage{makecell}%
\usepackage{threeparttable}%
\usepackage{caption}%
\usepackage[utf8]{inputenc}

\usepackage{microtype}

\usepackage{inconsolata}

\usepackage{graphicx}
\usepackage{enumitem}
\usepackage{xurl}%

\title{An Empirical Study of VLM Pipelines \\for Long-Document QA}

\author{Kenan E. Ak, Jay Mohta, Gwang Gook Lee, Yan Xu, Dimitrios Dimitriadis  \\
        Amazon.com \\
        \texttt{\{kenanea, jaymoht, gglee, yanxuml, dbdim\}@amazon.com}
        }

\begin{document}
\maketitle

\begin{abstract}
Vision-Language Models (VLMs) are increasingly used for long-document processing, where the inputs combine text with charts, tables, figures, and complex layouts. Deploying them means choosing how to feed the document to the model, which retriever to use when only a subset of pages is sent, and whether to run the model agentically or as a static pipeline. We study these choices on two long-document QA benchmarks with both frontier API and open-weight VLMs. First, on MMLongBench-Doc our six-tool agent with page, table, figure, and search calls pays off only once the answering VLM is large enough: with Qwen3.5-4B and 9B it trails static page input, with Qwen3.5-27B it draws level, and with Sonnet 4.5 it leads. On LongDocURL it is level with or ahead of static input at every reader. Its lead over the strongest static pipeline is clearest with the frontier reader on MMLongBench-Doc and narrows to within noise on LongDocURL. Second, retrieval modality matters more than the specific retriever: the strongest image retriever leads the strongest text pipeline, and on the text side a single off-the-shelf cross-encoder rerank essentially matches a much heavier multi-stage LLM pipeline. Top-$k$ image retrieval is also the most token-efficient input at every reader we paired it with, at roughly a seventh to a quarter of the tokens of sending every page.
Third, cutting across all three choices, three of our strongest pipelines succeed on different questions, and an oracle that picks the best pipeline per question gains roughly thirteen points over the best single pipeline, though evidence-type routing recovers almost none of it.
\end{abstract}

\section{Introduction}
Vision-Language Models (VLMs)~\cite{liu2023llava,wang2024qwen2vl,bai2025qwen3} are becoming the standard choice for long-document processing. Real-world long documents such as financial filings, official reports, and scientific papers span tens to hundreds of pages, interleaving prose with charts, tables, and figures under inconsistent layout. Answering a single question can require locating the right page, interpreting a specific chart or table, and combining evidence scattered across several pages. Short-context benchmarks~\cite{mathew2021docvqa,masry2022chartqa} rarely test these skills together.

Three architectural decisions shape a production long-document QA pipeline: the \emph{input modality} in which the document reaches the VLM, the \emph{retriever} that selects pages when only a subset is sent, and whether the model works \emph{agentically}, choosing pages, figures, or tables at inference time via tool calls rather than reading a fixed input (\S\ref{sec:task} defines all three). Acting agentically can surface evidence a retriever would miss, at the cost of latency and complexity. Each axis has been studied on its own, but their joint behaviour, and the accuracy/cost trade-offs it implies, have not been measured at frontier-VLM scale.

In this work, we present an empirical study of these three aspects on two long-document QA benchmarks (MMLongBench-Doc~\cite{ma2024mmlongbench} and LongDocURL~\cite{deng2024longdocurl}). We sweep several input modalities, retrieval approaches spanning lexical, dense-text, vision late-interaction, and rerank families, and several document-agent designs, with Claude Sonnet 4.5~\cite{anthropic2025claude45} served through a hosted API and Qwen3.5-\{4B, 9B, 27B\}~\cite{qwen2026qwen35} served locally. All configurations we run are scored on the full benchmark under one evaluation pipeline, producing a comparable leaderboard. Our six-tool function-calling agent with Sonnet 4.5 is the highest-scoring configuration under the official scorer on both benchmarks; on MMLongBench-Doc it leads same-reader top-5 retrieval by $+11.7$\,pp, and on LongDocURL it is level with the best retrieval pipeline.

Our contribution is not any single pipeline; the function-calling agent is a standard design. It is the controlled joint study of all three axes on one comparable footing, together with a re-evaluation of the closest prior system under each benchmark's official scorer rather than the more lenient LLM judges several such systems report against. That re-evaluation drops its judge-scored number by roughly $9$\,pp (\S\ref{sec:exp-agents}).

\section{Related Work}
\label{sec:related-work}

\subsection{Long-Document Multimodal QA}
\label{sec:related-work-benchmarks}

Document QA has scaled from single-page benchmarks~\cite{mathew2021docvqa,mathew2022infographicvqa,masry2022chartqa,tanaka2021visualmrc} to multi-page settings with moderate document length~\cite{tito2023mpdocvqa,tanaka2023slidevqa,vanlandeghem2023dude,dasigi2021qasper,lala2023paperqa}, and most recently to long, mixed-modality documents~\cite{chia2024mlongdoc,islam2023financebench,xia2024docgenome}. In this work, we utilize two benchmarks at the long end of this progression. MMLongBench-Doc~\cite{ma2024mmlongbench} comprises 135 PDFs with 1{,}082 questions stratified by single-page versus cross-page evidence and by visual-element type (chart, table, figure, layout, text). LongDocURL~\cite{deng2024longdocurl} comprises 396 long documents with 2{,}325 questions in its public split, organised along three tasks (Understanding, Reasoning, Locating) and four of the same evidence-element types (all but chart). We choose these two because they cover different parts of the space: MMLongBench-Doc is shorter but more visually varied, LongDocURL longer.

\subsection{Vision-Language Models}
\label{sec:related-work-vlms}

Vision-language models integrate a visual encoder with a language model so a single network reasons over text and images together~\cite{liu2023llava,qwen2026qwen35}. Document-specific VLMs instead encode page structure: layout-aware models fuse text, position, and visual features but require a separate OCR step~\cite{xu2020layoutlm}, OCR-free successors encode page images end-to-end~\cite{kim2022donut}, and high-resolution multi-page extensions compress each page's visual tokens to keep long inputs tractable~\cite{hu2024docowl2}. These are typically evaluated at smaller context rather than on documents of tens to hundreds of pages, where access mode and retrieval also matter.

Frontier multimodal models, on the API side~\cite{anthropic2025claude45,openai2024gpt4o,google2024gemini} and the open-weight side~\cite{qwen2026qwen35,chen2024internvl25,liu2024llavanext,wu2024deepseekvl2}, accept page images directly and increasingly do well on long-document QA without document-specific architectures. Long-context modeling has been studied mostly on the text side~\cite{liu2025comprehensive}, while evaluations on multimodal documents usually commit to a single access mode, whether all page images~\cite{ma2024mmlongbench,deng2024longdocurl}, a fixed set of retrieved pages~\cite{chia2024mlongdoc,han2025mdocagent}, or an agentic tool loop~\cite{lee2024readagent,jain2025simpledoc}, rather than comparing these choices head to head. This leaves the cost/accuracy trade-off across VLM scale largely open.

\subsection{Page Retrieval over Long Documents}
\label{sec:related-work-retrieval}

Lexical BM25~\cite{robertson2009bm25} over OCR or layout-aware text remains a strong and low-cost baseline for page retrieval. Dense bi-encoders~\cite{karpukhin2020dpr,chen2024bgem3} and late-interaction models~\cite{khattab2020colbert} match or exceed BM25 at higher compute cost, and ColPali~\cite{faysse2024colpali} extends this idea to vision by representing each page as an image. A second-stage rerank trades latency for accuracy, with cross-encoder~\cite{nogueira2019rerank} and LLM-based~\cite{sun2023rankgpt} variants commonly used. For long-document QA specifically, recent work either changes what gets indexed~\cite{jiang2024longrag} or layers LLM-driven steps such as document enrichment, query rewriting, and reranking on top of BM25~\cite{yang2026sira,gao2023hyde,asai2024selfrag}. The contribution of each step relative to the underlying text source is rarely isolated, which is what we examine in our retrieval ablation.

\subsection{Agentic and Tool-Using Document QA}
\label{sec:related-work-agents}

Tool-augmented agents~\cite{yao2023react,shinn2023reflexion} interleave tool calls with reasoning. For long documents, an agent explores the document on demand instead of reading a fixed input, trading one inference call for several tool-call rounds. Earlier text-only designs use gist memory or recursive document structure~\cite{lee2024readagent,chen2024memwalker,li2024graphreader}, and retrieval-augmented variants add search-and-fetch tools~\cite{bai2024qwenagent}.

Recent multimodal agents extend these designs to visual evidence. MDocAgent~\cite{han2025mdocagent} routes queries through five specialised agents on top of a multimodal RAG retriever, DocLens~\cite{zhu2025doclens} splits the task across navigation, localization, and reasoning roles, and SimpleDoc~\cite{jain2025simpledoc}, the closest analogue to our setup, runs an LLM reasoner that iteratively pulls retrieved pages into a working memory for a separate VLM answerer. Other concurrent designs train scrolling policies~\cite{lim2025scope}, decompose the task into planning, execution, judgement, and answer agents~\cite{yu2025mact}, or extend multimodal RAG to multi-document QA~\cite{suri2024visdom}. These systems span a wide compute range, from a single VLM with a small toolkit to multi-model stacks that pair a separate retrieval model with a 32B vision-language answerer, and most depend on per-document preprocessing such as per-page summaries or vision embeddings. They have largely been compared to other agentic systems rather than to a strong static retrieval-and-read pipeline.

\section{Method}
\label{sec:task}

We study long-document multimodal question answering as a deployment problem. A system $S$ is a function $S(D, q) \to a$ that maps a document $D$ (a long PDF with heterogeneous content) to a textual answer $a$ for a natural-language question $q$. The question may require evidence from a single page, non-adjacent pages, or none of the document at all. We decompose $S$ into three design choices that a practitioner can change one at a time. 
\begin{itemize}[nosep,leftmargin=*]
  \item \textbf{Input modality} $I$: what form of $D$ reaches the VLM that answers the question, which we call the \emph{reader} (raw PDF, page images, extracted text, or a retrieved subset).
  \item \textbf{Retriever} $R$: how pages are ranked by relevance to $q$ when $I$ keeps only the top $k$.
  \item \textbf{Agentic policy} $\pi$: trivial (single forward pass of static input) or a multi-turn tool-call loop in which the VLM decides what to look at next.
\end{itemize}

\paragraph{System instantiations.}
We evaluate five input modalities (raw PDF, all-pages images, plain-text, layout-aware text, and top-$k$ retrieved images), retrievers spanning lexical, dense-text, vision late-interaction, and rerank families, and several agent designs covering function-calling, gist-based, and iterative-refinement styles. Our readers are Claude Sonnet~4.5, served through a hosted API, and the open-weight Qwen3.5-\{4B, 9B, 27B\}, served locally. Appendices~\ref{sec:appendix-pdf}--\ref{sec:appendix-agents} give the full retriever and VLM configurations.

\paragraph{Evaluation protocol.}
All systems are scored on MMLongBench-Doc~\cite{ma2024mmlongbench} and LongDocURL~\cite{deng2024longdocurl}. Following each benchmark's own evaluation protocol, an \emph{answer extractor} LLM (Claude Haiku~4.5) normalises each free-form response into a short answer before comparison against the gold answer by the benchmark's official deterministic scorer; we use MMLongBench-Doc's released extraction prompt unchanged and apply it identically to every system we run, including the external baselines ReadAgent and SimpleDoc. It is distinct from the \emph{text} extractors (PyMuPDF, MinerU) used by the text-input modes. The step is not cosmetic: without it, correct answers inside explanatory prose score zero under the whole-string official scorers (Appendix~\ref{sec:appendix-significance}). We report per-question accuracy and mean input tokens, where the token count covers access-mode and agent-loop cost but excludes one-time preprocessing and retrieval-time compute, which Appendix~\ref{sec:appendix-retriever-cost} reports separately.

\paragraph{Per-question oracle.}
The \emph{per-question oracle} scores $\max_j \text{score}(S_j, q)$ over a fixed set of pipelines, crediting a question whenever \emph{any} member answers it. It is a hindsight upper bound rather than a deployable system, and its gap to the best single pipeline bounds what a per-question router could add. Unless stated otherwise it is taken over three of our strongest configurations; Appendix~\ref{sec:appendix-router} gives the member set and the fixed-membership caveat.

\paragraph{Experimental setup.}
Every VLM, including the answer extractor, decodes greedily (temperature${=}0$) with reasoning/thinking disabled, so all configurations we run in Table~\ref{tab:headline} share identical decoding; thinking mode is studied separately in \S\ref{sec:exp-input}. Page images render at DPI 144, the peak of a resolution sweep; all-images additionally caps each page at 1024\,px so that long documents fit in context (Appendix~\ref{sec:appendix-resolution}). Full per-pipeline configurations, image-size caps, generation limits, and determinism details are in Appendices~\ref{sec:appendix-resolution}--\ref{sec:appendix-agents}.

\subsection{Six-Tool Function-Calling Agent}
\label{sec:method-fc-agent}

Our agentic baseline is a function-calling loop in which a VLM explores the document on demand.

\paragraph{Preprocessing.}
We run MinerU~\cite{niu2025mineru25} on each PDF to extract its layout structure (section headings, figure/table crops, and per-page text blocks) and build a \emph{structural catalog}. The catalog lists, for each page, its section title, page range, and the figures/tables it contains, each with a stable integer id. This catalog is provided to the VLM as a compact preamble alongside the question.

\paragraph{Toolkit.}
The model chooses one or more tools at each turn from:
\begin{itemize}[nosep,leftmargin=*]
  \item \texttt{get\_catalog}: returns the full structural catalog (used when the preamble is truncated for very large documents).
  \item \texttt{read\_pages(pages)}: renders the requested pages as images at a model-chosen size, $700$ or $1{,}400$\,px on the longest side.
  \item \texttt{get\_figure(id)}: returns the MinerU-extracted figure crop by stable id.
  \item \texttt{get\_table(id)}: returns the table crop and its structured HTML.
  \item \texttt{search\_text(query, top\_k)}: runs BM25 over MinerU per-page text blocks and returns the top-scoring page numbers, each with a matching text excerpt.
  \item \texttt{finish(answer)}: commits the final answer and ends the loop.
\end{itemize}

The loop is capped at eight tool calls per question; in practice the median is two (Appendix~\ref{sec:appendix-tool-usage}). If a tool call references an invalid id, the agent receives an error message and may retry on the next turn.

\paragraph{Design rationale.}
Unlike gist-based agents~\cite{lee2024readagent}, ours sees actual page images, preserving visual evidence. Unlike multi-agent pipelines that fix retrieval mode up front~\cite{han2025mdocagent,jain2025simpledoc}, a single model can mix strategies within one question: for example, \texttt{search\_text} to locate a section, then \texttt{get\_table} to read a specific table. Dedicated element tools provide higher resolution than full-page rendering.

\section{Experiments}
\label{sec:experiments}

We organize results around the three architectural axes from Section \ref{sec:task}, namely input modality (\S\ref{sec:exp-input}), retrieval (\S\ref{sec:exp-retrieval}), and agentic access (\S\ref{sec:exp-agents}), studying each on both MMLongBench-Doc (MMLB) and LongDocURL (LDURL) and noting where findings transfer across the two benchmarks. Table~\ref{tab:headline} aggregates headline numbers across all configurations.

\begin{table}[t]
\centering
\scriptsize
\setlength{\tabcolsep}{3.5pt}
\renewcommand{\arraystretch}{0.88}
\begin{tabular}{@{}ll cc@{\hspace{4pt}} cc@{}}
\toprule
& & \multicolumn{2}{c}{MMLB} & \multicolumn{2}{c}{LDURL} \\
\cmidrule(lr){3-4} \cmidrule(lr){5-6}
Configuration & VLM & AVG & Tok & AVG & Tok \\
\midrule
Raw PDF                     & Sonnet  & 0.522 & 80k  & 0.414 & 150k \\
\midrule
All images                  & Sonnet  & 0.477 & 32k  & 0.360 & 44k \\
All images                  & Qwen3.5-4B & 0.526 & 33k  & 0.569 & 66k \\
All images                  & Qwen3.5-9B & 0.559 & 33k  & 0.592 & 66k \\
All images                  & Qwen3.5-27B & 0.588 & 30k & ---   & --- \\
\midrule
All text, PyMuPDF           & Sonnet  & 0.404 & 21k  & 0.595 & 51k \\
All text, MinerU            & Sonnet  & 0.480 & 27k  & 0.582 & 61k \\
All text, MinerU            & Qwen3.5-4B & 0.395 & 22k  & 0.353 & 41k \\
All text, MinerU            & Qwen3.5-9B & 0.390 & 22k  & 0.359 & 41k \\
All text, MinerU            & Qwen3.5-27B & 0.451 & 19k & 0.362 & 41k \\
\midrule
Top-5, BM25                 & Sonnet  & 0.407 & 7.5k & 0.555 & 7.4k \\
Top-5, ColQwen2.5           & Sonnet  & 0.508 & 7.6k & 0.643 & 7.5k \\
Top-5, BGE-M3+Coh.          & Sonnet  & 0.472 & 7.4k & 0.623 & 7.4k \\
Top-5, Nem-CE               & Sonnet  & 0.508 & 7.6k & 0.644 & 7.4k \\
Top-5, Nem-CE$\oplus$Scout  & Sonnet  & 0.533 & 7.6k & 0.615 & 7.5k \\
Top-5, Nem-CE$\oplus$Scout  & Qwen3.5-4B & 0.551 & 9.0k & 0.564 & 9.6k \\
Top-5, Nem-CE$\oplus$Scout  & Qwen3.5-9B & 0.564 & 9.0k & 0.576 & 9.6k \\
\midrule
Agent, ReadAgent            & Sonnet  & 0.498 & 11k & 0.612 & 21k \\
Agent, 6-tool FC            & Qwen3.5-4B & 0.471 & 77k & 0.598 & 70k \\
Agent, 6-tool FC            & Qwen3.5-9B & 0.497 & 33k & 0.605 & 36k \\
Agent, 6-tool FC            & Qwen3.5-27B & 0.616 & 33k & 0.644 & --- \\
Agent, 6-tool FC            & Sonnet   & \textbf{0.625} & 20k & \textbf{0.659} & 21k \\
\midrule
Agent, MDocAgent$^\dagger$ (5)  & Qwen2-VL-7B    & 0.315 & --- & 0.578 & --- \\
Agent, SimpleDoc$^\ddagger$ (2) & Qwen3+VL-32B   & 0.606 & --- & 0.723 & --- \\
Agent, SimpleDoc (2)            & Sonnet         & 0.608 & --- & 0.634 & --- \\
Agent, DocLens$^\S$ (3)         & Claude-4-S.    & 0.633 & --- & ---   & --- \\
Agent, DocLens$^\S$ (3)         & Gemini-2.5-Pro & 0.676 & --- & ---   & --- \\
\midrule
Oracle, 3-system & --- & 0.756 & --- & 0.788 & --- \\
\bottomrule
\end{tabular}
\caption{Headline leaderboard. \textbf{Configuration}: input mode, then retriever or agent. All images sends the whole document as page images, all text as extracted text; Top-5 sends five retrieved pages as images. AVG is each benchmark's official scorer, Tok the mean input tokens per question over the requests that completed; whole-document modes fail on the longest documents (raw PDF and Qwen text on LongDocURL, Qwen3.5-27B text on MMLongBench-Doc) and those questions score zero (Appendix~\ref{sec:appendix-pdf}). \textbf{Bold} marks the best result per benchmark among rows scored by the official scorer; our configurations are the 6-tool FC rows. Rows between the last two rules are external systems (agent count in parentheses); $^{\dagger}$/$^{\ddagger}$/$^{\S}$ mark numbers under the system's own protocol rather than ours ($\dagger$ GPT-4o, $\ddagger$ GPT-4.1 judge, $\S$ DocLens's own; \S\ref{sec:exp-agents}). Oracle is the per-question upper bound (\S\ref{sec:task}).}
\label{tab:headline}
\end{table}

\begin{figure}[t]
  \centering
  \includegraphics[width=0.9\linewidth]{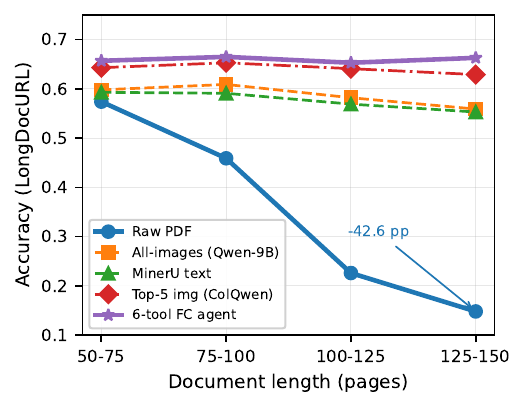}
  \caption{Accuracy vs.\ document page count, LongDocURL (Sonnet 4.5 unless a reader is named).}
  \label{fig:length}
\end{figure}

\begin{figure}[t]
  \centering
  \includegraphics[width=\linewidth]{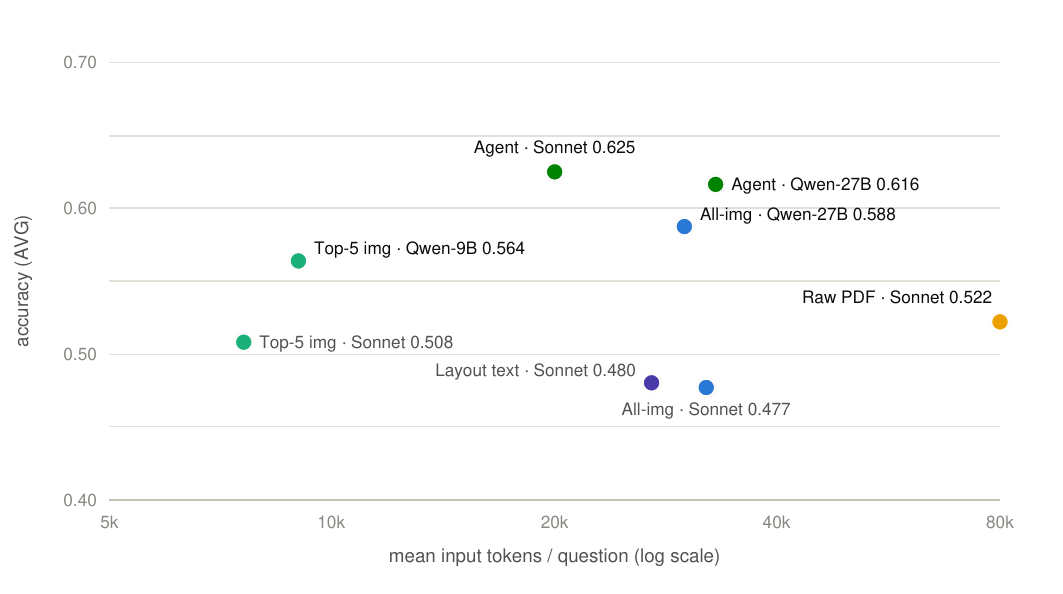}
  \caption{Accuracy vs.\ mean input tokens per question, MMLongBench-Doc (best
  configuration per access mode, plus the Sonnet 4.5 point where it differs;
  log-scale $x$, up-and-left better; no-think, official scorer).}
  \label{fig:tokens}
\end{figure}

\subsection{Input Modality}
\label{sec:exp-input}

We first compare the static whole-document modalities: raw PDF, all-images, and extracted text. Raw PDF is the strongest of these for Sonnet on MMLongBench (0.522) but also by far the most expensive, at 80k input tokens per question. The cost comes from how the API ingests a PDF, sending both a rendered image and an extracted text layer for every page. All-images is weaker than raw PDF in part because it sends the rendered pages without that text layer. It is also constrained by the API's 100-image-per-request limit: documents of more than 100 pages are packed into merged images at lower per-page resolution (Appendix~\ref{sec:appendix-pdf}). On MMLongBench, the Qwen3.5 family, served locally without this limit, beats Sonnet on the same mode by $5$--$11$\,pp, and by as much on documents of at most 50 pages, where Sonnet pays no penalty, so the gap is the reader, not the cap.

Text-only modes are the weakest input for the open-weight readers on MMLongBench, trailing all-images by $13$--$17$\,pp at 4B and 9B because they drop in-document images; for Sonnet they are competitive instead, since its all-images accuracy drops sharply on documents beyond about 50 pages. The strongest text-only configuration still trails top-$k$ image retrieval by about $5$\,pp at a matched reader, so reading the page beats reading its extracted text. Table questions are the one exception, where a layout-aware extractor recovers structured cell content and matches or beats images on MMLongBench. On MMLongBench, MinerU lifts Sonnet's text-only accuracy by about $8$\,pp over plain-text extraction with PyMuPDF, and the gain concentrates on table and figure questions. The benefit is benchmark-dependent, since on LongDocURL plain-text and layout-aware extraction tie at Sonnet. Appendix~\ref{sec:appendix-text-extractor} has the full sweep.

With text input, scale buys the Qwen3.5 family little below 27B. On MMLongBench the 4B and 9B models land within a point of each other, 8 to 9 pp below Sonnet on the same MinerU input; the 27B run fails on 111 long-document requests and scores zero there, and on the questions it completes it is about $11$\,pp above 9B. On LongDocURL the Qwen text runs fail outright on 895 of the 2,325 questions and score zero there, at a rate that climbs with document length from $18\%$ of the shortest bin to $72\%$ of the longest; on the questions they complete, the three sizes land within 2 pp of each other and of Sonnet.

Access mode also governs how gracefully accuracy scales with document length. Figure~\ref{fig:length} bins LongDocURL by page count: raw-PDF accuracy collapses by $42.6$\,pp from the 50--75 to the 125--150 page bin, mostly because the request itself fails above the API's per-request cap and scores zero, whereas retrieval, the agent, and all-images at a local reader that sees every page individually vary by at most $5$\,pp across the full range. The agent rows are the flattest, and the Sonnet agent is highest in every bin. The collapse is not an artifact of the single request: splitting documents into chunks under the cap and combining per-chunk answers recovers nothing (Appendix~\ref{sec:appendix-mapreduce}). Per-bin tables for both benchmarks are in Appendix~\ref{sec:appendix-page-count}.

Enabling a reasoning (``thinking'') budget before the model answers gives modest gains, up to about $6$\,pp and largest where the input is largest, and never changes the modality ordering; we report no-think numbers throughout for a matched-decoding comparison and defer the full sweep to Appendix~\ref{sec:appendix-thinking}.

Figure~\ref{fig:tokens} makes the accuracy/cost trade-off explicit, plotting each access mode's best configuration by accuracy against mean input tokens per question. Top-$k$ image retrieval sits in the efficient corner (0.51--0.56 at 7.6--9\,k tokens), the 6-tool agent is the most accurate (0.616--0.625 at 20--33\,k tokens), and raw PDF is the outlier, spending 80\,k tokens for 0.522. Only top-$k$ retrieval and the agent lie on the Pareto frontier: raw PDF and extracted text are dominated by both, and all-images (0.588 at 30\,k) is dominated by the Sonnet agent, which is both more accurate and cheaper (0.625 at 20\,k); top-$k$ is cheaper still but $6$\,pp less accurate than the agent.

\begin{table}[t]
\centering
\scriptsize
\setlength{\tabcolsep}{2pt}
\renewcommand{\arraystretch}{0.90}
\begin{tabular}{@{}llccc@{}}
\toprule
Retriever & Modality & MMLB & LDURL & Avg \\
\midrule
BM25 (MinerU)         & text       & 65.6 & 74.0 & 69.8 \\
BGE-M3                & text       & 70.8 & 77.4 & 74.1 \\
BM25 + Cohere 3.5     & text+rerank & 77.0 & 86.9 & 82.0 \\
BGE-M3 + Cohere 3.5   & text+rerank & 77.5 & 86.6 & 82.1 \\
SIRA (full pipeline)  & text+LLM   & 77.2 & 87.5 & 82.4 \\
ColQwen2.5            & image      & 84.2 & 90.1 & 87.2 \\
Nem-CE$\oplus$Scout   & image+LLM  & \textbf{91.3} & 82.5 & 86.9 \\
Nem-CE                & image      & 88.9 & \textbf{93.8} & \textbf{91.4} \\
\bottomrule
\end{tabular}
\caption{Hit@5 on the answerable subset; Avg is the cross-benchmark mean. \textbf{Bold} marks each column's best. Per-retriever cost details are in Appendix~\ref{sec:appendix-retriever-cost}.}
\label{tab:retrieval-hit5}
\end{table}

\subsection{Retrieval}
\label{sec:exp-retrieval}

A retrieval pipeline ranks the document's pages and sends only the top $k$ to the VLM. We use $k=5$ throughout unless stated otherwise, the depth of every Top-5 row in Table~\ref{tab:headline}; a depth sweep (Appendix~\ref{sec:appendix-ksweep}) shows accuracy peaks near $k{=}10$ and that $k{=}5$ captures most of it at roughly half the tokens. We measure retrieval with Hit@5, the fraction of answerable questions whose top-5 pages include an evidence page. We evaluate retrievers from four families: BM25~\cite{robertson2009bm25} (lexical word overlap), BGE-M3~\cite{chen2024bgem3} (dense-text bi-encoder), ColQwen2.5~\cite{faysse2024colpali} and Nemotron-ColEmbed-3B (Nem-CE)~\cite{moreira2026_nemotron_colembed_v2} (late-interaction over page images), and rerank pipelines that stack a cross-encoder or full LLM rerank on a base retriever, including SIRA~\cite{yang2026sira}. We also build a cascade: Nem-CE-3B first prefilters each document to 50 candidate pages, and Scout, a Qwen3.5-9B model prompted to read those candidates and rank them, selects the final top 5.

As shown in Table~\ref{tab:retrieval-hit5}, modality is the dominant axis in retrieval. Averaged across both benchmarks, the image retrievers ColQwen2.5 and Nem-CE reach 87.2 and 91.4 Hit@5, while no text retriever exceeds 82.4, so the strongest image retriever leads the strongest text pipeline by about 9\,pp and the strongest base image retriever (Nem-CE) leads the strongest base text retriever (BGE-M3) by roughly 17\,pp. The effect carries downstream to answer accuracy, not just retrieval: with the same reader, five retrieved page images match the whole document's extracted text on MMLongBench (0.508 vs.\ 0.480, a paired gap whose interval includes zero) while sending under a third of the input tokens (7.6k vs.\ 27k), and the per-evidence breakdown favours images on every source except Tables, where MinerU's structured cell HTML wins on MMLongBench though not on LongDocURL (Table~\ref{tab:per-evidence}).

On the text side, the reranker matters more than the base retriever. Dense BGE-M3 leads lexical BM25 by only $4.3$\,pp on average (74.1 vs.\ 69.8), but adding Cohere Rerank 3.5~\cite{cohere2024rerank} lifts both by 8 to 12\,pp and renders them indistinguishable afterward (82.0 vs.\ 82.1). A single such rerank is statistically indistinguishable from SIRA's full multi-stage pipeline (82.4 vs.\ 82.0; on MMLongBench the paired Hit@5 difference is $+0.1$\,pp, $95\%$ CI $[-2.8,+2.9]$, Appendix~\ref{sec:appendix-significance}), with no LLM in the loop (Appendix~\ref{sec:appendix-retriever-cost}).

Reranking does not carry over to the already-strong image retrievers. Stacking the Scout reranker on Nem-CE gives the best MMLongBench Hit@5 in the table ($+2.4$\,pp) but loses $11.3$\,pp on LongDocURL, a net drop of $4.5$\,pp on average against plain Nem-CE. The loss is a document-length effect: binning LongDocURL by page count, Scout's Hit@5 deficit against plain Nem-CE widens from $-5.7$\,pp on 50--75-page documents to $-22.0$\,pp on 125+-page documents, while Nem-CE alone stays flat ($0.92$--$0.95$ across all bins). On long documents Scout increasingly demotes the evidence page that Nem-CE alone had ranked in its top five; MMLongBench's shorter documents rarely trigger this, which is why the cascade helps there. The full per-bin breakdown is in Appendix~\ref{sec:appendix-scout}.

\subsection{Agentic Access}
\label{sec:exp-agents}

As shown in Table~\ref{tab:headline}, the 6-tool function-calling (FC) agent is the highest-scoring single-VLM configuration we measured at frontier scale on both benchmarks; it beats the best retrieve-then-generate pipeline on MMLongBench and is level with it on LongDocURL.

This advantage appears only above a scale threshold. On MMLongBench the FC agent \emph{trails} both static image modes at Qwen3.5-4B (0.471 vs.\ 0.526 all-images and 0.551 top-5) and at 9B (0.497 vs.\ 0.559 and 0.564), draws level with the best of them at 27B (0.616 vs.\ 0.588, a $+2.8$\,pp paired gap whose interval includes zero), and leads at Sonnet 4.5 (0.625 vs.\ 0.533). The transition sits between 9B and 27B: below it, a reader is better served by simply being shown every page. Among static modes at open-weight scale, top-$k$ retrieval and all-images are indistinguishable in accuracy, with the paired interval including zero at both 4B and 9B, while top-$k$ uses under a third of the input tokens (9k vs.\ 33k at 9B), so top-$k$ is the better static default rather than merely the cheaper one. LongDocURL shows no such threshold: the agent leads top-$k$ retrieval at its own reader already at 4B and 9B, and at 27B and Sonnet 4.5 it is level with the best static pipeline; the largest step is again 9B to 27B. The smallest agent is also the most expensive, and not because it searches more: every Qwen reader reaches its answer within two to three tool calls, and what separates them is whether they stop. Qwen3.5-4B re-emits \texttt{finish} after its answer is recorded on $51\%$ of MMLongBench trajectories and $75\%$ on LongDocURL, adding $9.1$ and $4.9$ further model calls that re-send the accumulated context, against $1.5$ and $0.3$ for 27B. Knowing when to stop is a capability: toolkit and loop are identical across the Qwen readers, so 4B's $77$k against $33$k is the price of a reader that does not end its loop (Appendix~\ref{sec:appendix-agents}).

ReadAgent~\cite{lee2024readagent}, which uses a gist-based design, lands well below our toolkit on MMLongBench and only modestly below on LongDocURL. It looks up a median of two pages per question and reads them through a compressed gist memory, so it often misses the evidence page and loses the fine-grained list and table detail that our structured catalog preserves. The gains therefore come from the toolkit, not from acting agentically alone.

\paragraph{Comparison with multi-agent and iterative-refinement methods.}
We evaluate three multi-agent systems on the same two benchmarks. MDocAgent~\cite{han2025mdocagent} routes a question through five agents and answers with Qwen2-VL-7B over ColPali top-$k=4$ pages. SimpleDoc~\cite{jain2025simpledoc} runs a Qwen3-30B-A3B reasoner that repeatedly pulls dual-cue retrieved pages into a working memory before a Qwen2.5-VL-32B answerer responds. DocLens~\cite{zhu2025doclens} splits the task across a Page Navigator, an Element Localizer, and a Reasoner.

MDocAgent and SimpleDoc score under binary LLM judges (GPT-4o and GPT-4.1 respectively) rather than each benchmark's official deterministic scorer, so the published figures sit above what the official scorer would assign. To place SimpleDoc on the same footing, we re-ran its full pipeline with Sonnet 4.5 in every role under the deterministic scorer. This reproduces its MMLongBench result (0.608 vs.\ 0.606) and yields 0.634 on LongDocURL, about $8.9$\,pp below the judge-scored 0.723.

The 6-tool FC agent is competitive with all of these systems, but only the SimpleDoc comparison is strictly like-for-like. At a matched VLM under the same deterministic scorer our agent is $1.6$\,pp ahead on MMLongBench, $[-1.1,+4.4]$, which we report as a tie. The DocLens rows use its own protocol, so those comparisons are indicative; our closest row is the Sonnet agent, level with DocLens's Claude-4-Sonnet configuration and below its Gemini-2.5-Pro one. MDocAgent reaches only 0.315, far enough back that the judge leniency cannot account for the gap. On LongDocURL the same agent leads SimpleDoc-Sonnet by $2.5$\,pp and MDocAgent by about $8$\,pp.

A single strong VLM with a structural catalog and direct figure and table access recovers visual evidence more reliably than a pipeline of smaller specialized models, and at lower cost: our agent runs one VLM over a lightweight MinerU-derived catalog in a median of two tool calls, with no second large model in the loop and no per-page offline indexing. Appendix~\ref{sec:appendix-agent-cost} sets all five agentic systems side by side on this cost axis.

\section{Conclusion}
\label{sec:conclusion}
We studied three deployment-level choices for long-document multimodal QA (input modality, retrieval, and agentic access) across two benchmarks under one extractor and evaluator. The findings reduce to a decision tree. \emph{With a frontier API}, a six-tool function-calling agent is the highest-accuracy configuration we measured, at roughly $1.5$--$7\times$ fewer input tokens per question than sending all pages as images or the raw PDF. \emph{With a smaller open-weight VLM} (Qwen3.5-9B and below), tool use is a poor trade: on MMLongBench the agent trails plain all-images input, on LongDocURL its lead over top-5 retrieval is a few points, and in both cases it spends $4$--$9\times$ the tokens of top-5 retrieval. Static image input is the better default at this scale; top-5 retrieval matches all-images on accuracy at a fraction of the tokens. \emph{When images cannot be sent}, a layout-aware extractor (MinerU) with BM25 or BGE-M3 and a single cross-encoder rerank matches a multi-stage design's text-side accuracy at a fraction of its compute. \emph{Beyond any single pipeline}, three of our strongest configurations make different mistakes and leave roughly $13$\,pp of per-question oracle headroom (Table~\ref{tab:headline}), but that headroom resists the obvious remedy: routing on gold evidence type recovers under a point of it, and a router on observable question features is \emph{worse} than always running the agent ($-0.9$\,pp on MMLongBench, $-7.1$\,pp on LongDocURL; Appendix~\ref{sec:appendix-router}), because the complementarity lies within evidence types rather than across them, and recovering it needs a finer per-question signal than any we tested. A small, well-chosen toolkit can stand in for a larger multi-agent stack.

\section*{Limitations}

A few aspects of our setup limit the scope of our findings. We study long-document \emph{question answering} only, the task both benchmarks target, and do not measure summarization, extraction, or classification; while the three axes we vary (input modality, retrieval, and agentic access) are task-general choices, whether the specific rankings transfer to other long-document tasks is left to future work, and we do not claim generality beyond QA. The two benchmarks together contain 3{,}407 questions over 531 long documents, and differences below about $3$\,pp on MMLongBench-Doc and $2$\,pp on LongDocURL are within noise (Appendix~\ref{sec:appendix-significance}), so we avoid fine-grained ordering claims at that scale. The unified scorer combines an answer-extractor LLM with a benchmark-native string-and-numeric matcher, and when a system already returns a short structured answer (such as the function-calling agent's \texttt{finish} argument), the extractor adds little, which may slightly understate that system's lead. On the API side, Sonnet caps the number of images per request and limits each image to 8000 pixels on a side, and our pipeline silently falls back by stitching and resizing page images for very long documents; beyond a further size threshold the request fails outright and is scored zero (Appendix~\ref{sec:appendix-pdf}). Numbers we report from commercial VLM endpoints reflect that fallback, and on-premise serving may behave differently. Finally, our results assume English-language documents and the VLM families we tested, and whether the rankings carry over to other languages and to smaller on-device VLMs is open. Errors concentrate on charts and figures on MMLongBench-Doc and on layout-bearing pages on LongDocURL (Table~\ref{tab:per-evidence}), so a benchmark with a different evidence mix would likely reorder the leaderboard.

\section*{Ethical Considerations}

This work uses two public document QA benchmarks (MMLongBench-Doc and LongDocURL), neither of which contains personally identifying information or sensitive personal data. All VLMs and retrievers are publicly released models or commercial APIs accessed under their standard terms of service. We do not collect any new human-labeled data and run no experiments with human subjects. The systems we evaluate produce textual answers grounded in the input document, which limits the risk of fabrication, but a deployed system should still surface its source pages to the user so that incorrect answers are auditable. Our cost numbers are reported in input tokens per question rather than in dollars to avoid pinning recommendations to any single provider's pricing, which can change.

\bibliography{references}

\appendix

\section{Per-Evidence-Source Accuracy}
\label{sec:appendix-per-evidence}

Table~\ref{tab:per-evidence} reports per-question accuracy on the subset of questions tagged with each evidence-source label. Questions with multiple labels appear in each. MMLongBench-Doc tags are TXT (Pure-text), LAY (Generalized-text/Layout), CHA (Chart), TAB (Table), FIG (Figure). LongDocURL uses four of these, all but CHA.

\begin{table*}[t]
\centering
\scriptsize
\setlength{\tabcolsep}{3pt}
\begin{tabular}{@{}lllc ccccc c cccc@{}}
\toprule
& & & & \multicolumn{5}{c}{MMLongBench-Doc} & & \multicolumn{4}{c}{LongDocURL} \\
\cmidrule(lr){5-9} \cmidrule(lr){11-14}
Access & Retriever / Agent & VLM & $k$ & TXT & LAY & CHA & TAB & FIG & & TXT & LAY & FIG & TAB \\
\midrule
Raw PDF       & ---                  & Sonnet 4.5  & --- & 0.531 & 0.555 & 0.516 & 0.625 & 0.432 & & 0.443 & 0.367 & 0.419 & 0.378 \\
\midrule
All-pages img & ---                  & Sonnet 4.5  & --- & 0.495 & 0.513 & 0.486 & 0.515 & 0.456 & & 0.381 & 0.326 & 0.389 & 0.329 \\
All-pages img & ---                  & Qwen3.5-4B  & --- & 0.459 & 0.515 & 0.470 & 0.493 & 0.409 & & 0.636 & 0.503 & 0.548 & 0.553 \\
All-pages img & ---                  & Qwen3.5-9B  & --- & 0.482 & 0.496 & 0.531 & 0.538 & 0.463 & & 0.639 & 0.502 & 0.627 & 0.589 \\
\midrule
PyMuPDF text  & PyMuPDF              & Sonnet 4.5  & --- & 0.416 & 0.361 & 0.319 & 0.507 & 0.196 & & 0.661 & 0.516 & 0.589 & 0.583 \\
MinerU text   & MinerU               & Sonnet 4.5  & --- & 0.488 & 0.422 & 0.253 & 0.616 & 0.286 & & 0.674 & 0.516 & 0.525 & 0.533 \\
MinerU text   & MinerU               & Qwen3.5-4B  & --- & 0.421 & 0.347 & 0.173 & 0.467 & 0.236 & & 0.437 & 0.378 & 0.323 & 0.232 \\
MinerU text   & MinerU               & Qwen3.5-9B  & --- & 0.411 & 0.294 & 0.173 & 0.434 & 0.223 & & 0.442 & 0.377 & 0.328 & 0.238 \\
MinerU text   & MinerU               & Qwen3.5-27B & --- & 0.469 & 0.423 & 0.194 & 0.438 & 0.252 & & 0.441 & 0.389 & 0.340 & 0.246 \\
\midrule
Top-$k$ img   & BM25                 & Sonnet 4.5  & 5   & 0.372 & 0.348 & 0.425 & 0.414 & 0.267 & & 0.645 & 0.465 & 0.590 & 0.517 \\
Top-$k$ img   & ColQwen2.5           & Sonnet 4.5  & 5   & 0.535 & 0.533 & 0.511 & 0.487 & 0.481 & & 0.695 & 0.543 & 0.682 & 0.653 \\
Top-$k$ img   & BGE-M3$\oplus$Cohere & Sonnet 4.5  & 5   & 0.475 & 0.407 & 0.464 & 0.534 & 0.361 & & 0.706 & 0.539 & 0.645 & 0.608 \\
Top-$k$ img   & Nem-CE               & Sonnet 4.5  & 5   & 0.517 & 0.516 & 0.515 & 0.558 & 0.457 & & 0.697 & 0.554 & 0.670 & 0.655 \\
Top-$k$ img   & Nem-CE$\oplus$Scout & Sonnet 4.5  & 5   & 0.537 & 0.515 & 0.521 & 0.650 & 0.467 & & 0.666 & 0.534 & 0.632 & 0.610 \\
Top-$k$ img   & Nem-CE$\oplus$Scout & Qwen3.5-4B  & 5   & 0.501 & 0.530 & 0.509 & 0.547 & 0.445 & & 0.595 & 0.506 & 0.551 & 0.566 \\
Top-$k$ img   & Nem-CE$\oplus$Scout & Qwen3.5-9B  & 5   & 0.503 & 0.474 & 0.532 & 0.590 & 0.448 & & 0.601 & 0.499 & 0.593 & 0.580 \\
\midrule
Agent & ReadAgent          & Sonnet 4.5  & 7   & 0.493 & 0.445 & 0.497 & 0.561 & 0.379 & & 0.686 & 0.511 & 0.637 & 0.599 \\
Agent & 6-tool FC          & Qwen3.5-4B  & --- & 0.421 & 0.362 & 0.310 & 0.532 & 0.314 & & 0.661 & 0.497 & 0.567 & 0.577 \\
Agent & 6-tool FC          & Qwen3.5-9B  & --- & 0.438 & 0.379 & 0.370 & 0.558 & 0.411 & & 0.661 & 0.513 & 0.578 & 0.593 \\
\textbf{Agent} & \textbf{6-tool FC} & \textbf{Qwen3.5-27B} & --- & 0.578 & 0.545 & 0.557 & 0.706 & 0.524 & & 0.674 & 0.568 & 0.614 & 0.659 \\
\textbf{Agent} & \textbf{6-tool FC} & \textbf{Sonnet 4.5}  & --- & 0.606 & 0.578 & 0.526 & 0.701 & 0.530 & & 0.722 & 0.566 & 0.659 & 0.634 \\
Multi-agent$^\dagger$ & MDocAgent & Qwen2-VL-7B & 4 & 0.401 & 0.294 & 0.347 & 0.323 & 0.321 & & --- & --- & --- & --- \\
Single-agent$^\ddagger$ & SimpleDoc & Qwen3-30B+VL-32B & 3.5 & 0.599 & 0.513 & 0.549 & 0.512 & 0.512 & & --- & --- & --- & --- \\
Single-agent & SimpleDoc & Sonnet 4.5      & 3.5 & 0.583 & 0.665 & 0.576 & 0.654 & 0.539 & & 0.693 & 0.537 & 0.655 & 0.636 \\
Multi-agent & DocLens & Claude-4-Sonnet & --- & 0.599 & 0.582 & 0.544 & 0.639 & 0.553 & & --- & --- & --- & --- \\
Multi-agent & DocLens & Gemini-2.5-Pro & --- & 0.637 & 0.646 & 0.643 & 0.697 & 0.602 & & --- & --- & --- & --- \\
\bottomrule
\end{tabular}
\caption{Per-evidence-source accuracy on both benchmarks, broken out from the AVG columns of Table~\ref{tab:headline}. \textbf{Bold} marks our two strongest configurations. As in Table~\ref{tab:headline}, $^{\dagger}$ and $^{\ddagger}$ mark scores under the system's own LLM judge ($\dagger$ GPT-4o, $\ddagger$ GPT-4.1) rather than the official scorer. The configurations are the same; this table has room to split the input mode from the retriever or agent and to name both in full, where Table~\ref{tab:headline} combines them into one Configuration cell. ``---'' marks not applicable or not logged.}
\label{tab:per-evidence}
\end{table*}

\section{How the Claude API Processes Raw PDFs}
\label{sec:appendix-pdf}

The raw-PDF row in Table~\ref{tab:headline} hands the unmodified PDF to the Claude API rather than rendering page images ourselves, so its behaviour is set by the API's own document pipeline. We access Claude through its hosted API, and we document its handling here because it explains both the high token cost of raw-PDF input and its collapse on long documents (Appendix~\ref{sec:appendix-page-count}).

\paragraph{Page-as-image-plus-text.}
With citations enabled, the Claude API converts each PDF page into an image and provides the text extracted from that page alongside it, so the model sees both a rendered view and an OCR-style text layer for every page, at roughly $7$\,k tokens for a three-page PDF. Without citations it falls back to basic text extraction at roughly $1$\,k tokens for the same pages and cannot analyse charts, figures, or visual layout. We enable citations in every raw-PDF run so that Claude sees the visual content, which is why the raw-PDF token counts in Table~\ref{tab:headline} are the highest of any access mode.

\paragraph{Per-page cost and request limits.}
Under full visual mode each page costs roughly $1.5$\,k to $3$\,k text tokens depending on content density, plus the image tokens for the rendered page. A single request is capped at $32$\,MB and at $100$ pages for the $200$\,k-token-context models we use. When a document exceeds these limits the API downsamples the embedded page images to fit, which reduces per-page resolution on long documents; beyond a further threshold the request fails outright and we score it zero, which accounts for most of the raw-PDF accuracy collapse reported in Appendix~\ref{sec:appendix-page-count}. Failures concentrate in the two modes that send an entire long document in a single call: $736$ of $2{,}325$ LongDocURL questions under raw PDF and $895$ under MinerU text with the Qwen readers, against at most $38$ in every other LongDocURL configuration. On MMLongBench-Doc the only such failures are $111$ of $1{,}082$ questions under MinerU text with Qwen3.5-27B, all on documents of $72$ pages or more.

We confirm this downsampling directly from the logged token counts rather than inferring it. On LongDocURL raw-PDF runs, the mean input tokens charged \emph{per page} fall from ${\sim}2{,}050$ on $50$--$100$-page documents (mean $72$ pages) to ${\sim}1{,}540$ on $100$--$150$-page documents (mean $121$ pages). The effect is far larger for all-images, which carries no text layer to hold the count up: on LongDocURL its per-page cost falls from ${\sim}914$ to ${\sim}60$ tokens across the same two bins, a ${\sim}15\times$ reduction that sets in as documents cross the $100$-image request cap, and on MMLongBench-Doc it declines monotonically with length ($999 \to 614 \to 596$ over the $<\!50$, $50$--$100$, $100$--$150$ page bins). The collapse is therefore a measured consequence of the API packing more pages into a fixed budget, not an artifact we assume.

\section{Text-Extractor Sweep for Plain-Text Input}
\label{sec:appendix-text-extractor}

The plain-text input row reported in Table~\ref{tab:headline} fixes the extractor to PyMuPDF and the VLM to Sonnet 4.5. Table~\ref{tab:b1-extractor-sweep} shows the full extractor $\times$ VLM sweep. Two patterns hold across VLM scales. First, PyMuPDF and pdfplumber land within $1.5$\,pp of each other on the same VLM, so the choice of plain-text extractor does not matter. Second, at frontier VLM scale (Sonnet) the gap between either plain-text extractor and MinerU's layout-aware extractor (the MinerU text rows in Table~\ref{tab:headline}) is $+8$\,pp, and that gap remains $+2$ to $+5$\,pp at smaller open-weight scales. The text-extractor choice therefore matters less than the modality choice itself.

\begin{table}[h]
\centering
\scriptsize
\setlength{\tabcolsep}{4pt}
\begin{tabular}{@{}llcc@{}}
\toprule
Extractor & VLM & AVG & Tok-in \\
\midrule
PyMuPDF    & Qwen3.5-4B & 0.355 & 17.4k \\
PyMuPDF    & Qwen3.5-9B & 0.371 & 17.4k \\
pdfplumber & Sonnet 4.5 & 0.401 & 22.7k \\
pdfplumber & Qwen3.5-4B & 0.345 & 16.9k \\
pdfplumber & Qwen3.5-9B & 0.357 & 16.9k \\
\bottomrule
\end{tabular}
\caption{Plain-text PDF text-extractor sweep on MMLongBench-Doc. The PyMuPDF/Sonnet row in Table~\ref{tab:headline} is the canonical plain-text number, and this table reports the remaining cells we collected.}
\label{tab:b1-extractor-sweep}
\end{table}

\section{Resolution Sweep for Top-$k$ Retrieved Images}
\label{sec:appendix-resolution}

\paragraph{Generation limit and image-size caps.}
The generation cap is effectively non-binding: non-thinking answers average
${\sim}160$ tokens. The all-images mode caps each page at 1024\,px so that long
documents fit in context (${\sim}760$ vs.\ ${\sim}1{,}800$ tokens per uncapped
page), while the top-$k$ modes send five uncapped pages; this is why all-images
and top-$k$ token counts in Table~\ref{tab:headline} are not directly comparable.

For the top-$k$ retrieved-images setting we ran a fixed-DPI sweep on Qwen3.5-4B and Sonnet 4.5. Both peak at DPI=144. On Sonnet, input-token cost is flat for DPI~$\geq 144$ because the API tiles every page into a fixed grid of image tokens, so source pixels above $\sim$DPI=144 are internally discarded. Higher DPI on commercial endpoints is therefore a strict regression, with more JPEG/resize artifacts and no extra model-visible detail. On Qwen3.5-4B the picture is similar. Accuracy still saturates at DPI=144 (0.541, 0.557, 0.551, 0.548 at DPI 96, 144, 180, 216), but raising DPI keeps inflating input tokens 2--3$\times$ for $\leq 1$\,pp accuracy. The smart-resize adaptive cap at $1024\times1024$ pixels underperforms fixed DPI on both VLMs. Sonnet's measured curve on MMLongBench is given in Table~\ref{tab:sonnet-dpi}, and the mean per-question input-token count is essentially flat between DPI 144 and 216 ($7{,}535$--$7{,}579$\,tok) while accuracy peaks at 144 and degrades at higher DPI by $-1.8$ to $-2.0$\,pp.

\begin{table}[t]
\centering
\scriptsize
\setlength{\tabcolsep}{6pt}
\begin{tabular}{@{}lrrr@{}}
\toprule
DPI & AVG & Mean tok-in & $\Delta$ vs.\ 144 \\
\midrule
96   & 0.482 & 5{,}379 & $-2.6$\,pp \\
\textbf{144} & \textbf{0.508} & \textbf{7{,}579} & \textbf{0.0\,pp} \\
180  & 0.490 & 7{,}535 & $-1.8$\,pp \\
216  & 0.488 & 7{,}548 & $-2.0$\,pp \\
\bottomrule
\end{tabular}
\caption{Sonnet 4.5 top-5 retrieved images (ColQwen2.5) accuracy and mean per-question input tokens on MMLongBench-Doc at four fixed DPIs. Above DPI=144, mean input tokens are flat ($7{,}535$--$7{,}579$\,tok) because the API internally tiles each page into a fixed grid of image tokens. Raising DPI above 144 supplies more source pixels but no additional model-visible tokens, and degrades accuracy through extra JPEG/resize artifacts.}
\label{tab:sonnet-dpi}
\end{table}

We also sweep an explicit per-image pixel cap (\texttt{max\_image\_size}), which bounds the longest side of each rendered page rather than its render density. Table~\ref{tab:sonnet-cap} reports Sonnet 4.5 on the same top-5 ColQwen2.5 setting. Accuracy rises monotonically with the cap and saturates near 2048\,px, recovering the DPI=144 result (0.501 vs 0.508), while a 512\,px cap is far too low and destroys small-text and chart detail. The pixel cap is thus a second resolution control that behaves like DPI, with no setting beating the DPI=144 baseline.

\begin{table}[t]
\centering
\scriptsize
\setlength{\tabcolsep}{6pt}
\begin{tabular}{@{}lrrr@{}}
\toprule
Cap (px) & AVG & Mean tok-in & $\Delta$ vs.\ DPI 144 \\
\midrule
512  & 0.296 & 1{,}378 & $-21.2$\,pp \\
1024 & 0.484 & 5{,}099 & $-2.4$\,pp \\
\textbf{2048} & \textbf{0.501} & 7{,}578 & $-0.7$\,pp \\
\bottomrule
\end{tabular}
\caption{Sonnet 4.5 top-5 retrieved images (ColQwen2.5) on MMLongBench-Doc under a per-image pixel cap (\texttt{max\_image\_size}). Accuracy climbs with the cap and saturates near 2048\,px, matching the DPI=144 baseline (0.508) within noise; a 512\,px cap collapses accuracy by destroying fine detail. $\Delta$ is relative to the DPI=144 row of Table~\ref{tab:sonnet-dpi}.}
\label{tab:sonnet-cap}
\end{table}

\section{Scaling with Document Length}
\label{sec:appendix-page-count}

LongDocURL spans 51--149 pages per document, while MMLongBench-Doc spans 9--468 pages with a mean of 48, so LongDocURL lets us isolate how each access mode degrades as page count grows over a tighter, longer-skewed range. We bin LongDocURL questions by their document's total page count and compute Sonnet accuracy per bin under the official LongDocURL scorer (Table~\ref{tab:page-count-bins}). Two patterns are clear.

First, raw-PDF access collapses on long documents, from $0.574$ on 50--75-page documents to $0.148$ on 125--150-page documents, a $-42.6$\,pp drop. Most of that drop is outright request failure rather than degraded reading: above the API's per-request size cap the call returns no answer and scores zero, and the failure rate climbs from $8\%$ in the shortest bin to $68\%$ in the longest. Among the requests that do complete, accuracy falls far more gently, from $0.624$ to $0.467$. Second, every other mode (all-pages images, retrieved images, MinerU text, and the agent) is essentially flat across the four bins, with drops in the range of $-4$ to $+1$\,pp. Selective access modes therefore benefit not only in absolute accuracy but also in stability as document length scales, and the practitioner-facing implication is that any deployment expecting documents above ${\sim}80$ pages should avoid raw-PDF input regardless of VLM capability.

\begin{table}[t]
\centering
\scriptsize
\setlength{\tabcolsep}{4pt}
\resizebox{\columnwidth}{!}{
\begin{tabular}{@{}lrrrrr@{}}
\toprule
Pipeline                      & Overall & 50--75 & 75--100 & 100--125 & 125--150 \\
\midrule
Raw PDF                            & 0.414  & 0.574 & 0.459 & 0.226 & 0.148 \\
MinerU text                        & 0.582  & 0.593 & 0.591 & 0.569 & 0.553 \\
Top-5 img (BM25)                   & 0.555  & 0.540 & 0.576 & 0.573 & 0.531 \\
Top-5 img (ColQwen2.5)             & 0.643  & 0.643 & 0.653 & 0.641 & 0.629 \\
All-pages img (Qwen3.5-9B)         & 0.592  & 0.598 & 0.609 & 0.582 & 0.559 \\
\textbf{6-tool FC (Sonnet)}         & \textbf{0.659}  & \textbf{0.657} & \textbf{0.665} & \textbf{0.653} & \textbf{0.663} \\
6-tool FC (Qwen3.5-27B)               & 0.644  & 0.646 & 0.645 & 0.645 & 0.636 \\
\bottomrule
\end{tabular}
}
\caption{LongDocURL accuracy by document page-count bin (under the official LongDocURL scorer). \textbf{Bold} marks the best pipeline in each column. Rows use Sonnet 4.5 unless a reader is named. Each bin reports mean accuracy on questions whose document has total pages in the listed range. Raw PDF drops $42.6$\,pp from 50--75 to 125--150 pages, while every other mode changes by at most $4$\,pp between those bins and varies by at most $5$\,pp across all four.}
\label{tab:page-count-bins}
\end{table}

\paragraph{Cross-benchmark check on MMLongBench-Doc.}
We repeat the same per-bin analysis on MMLongBench-Doc (135 PDFs, 9--468 pages) to test whether the page-count effect is benchmark-specific. The bin distribution is heavily skewed (99 docs in 11--50 pages, only 14 in 101--200 and 1 above 200), so the longest-bin estimates are noisier. Patterns across the dense 11--50 vs 101--200 bins still tell a consistent story (Table~\ref{tab:mmlb-page-count}). Raw PDF drops only $-6.3$\,pp from $0.542$ to $0.479$, an order of magnitude milder than LDURL's collapse, because fewer MMLB documents cross the API's auto-stitch threshold. Top-$k$ retrieval with BM25 and ColQwen2.5 drops about $11$\,pp, while the Nem-CE$\oplus$Scout cascade barely moves ($-0.8$\,pp) and the 6-tool FC agent is similarly stable, dropping only a few points across the dense bins (Sonnet: $0.644 \to 0.591$, Qwen3.5-27B: $0.633 \to 0.611$), reproducing the catalog-gated stability we see on LongDocURL.

\begin{table}[t]
\centering
\scriptsize
\setlength{\tabcolsep}{4pt}
\resizebox{\columnwidth}{!}{
\begin{tabular}{@{}lrrrrr@{}}
\toprule
Pipeline                       & $\le$10 & 11--50 & 51--100 & 101--200 & $>$200 \\
\midrule
Raw PDF (Sonnet)                  & 0.370 & 0.542 & 0.477 & 0.479 & 0.400 \\
All-pages img (Qwen3.5-9B)        & 0.375 & 0.605 & 0.459 & 0.429 & 0.400 \\
Top-5 img (BM25, Sonnet)          & 0.307 & 0.429 & 0.366 & 0.316 & 0.800 \\
Top-5 img (ColQwen2.5, Sonnet)    & 0.495 & 0.539 & 0.414 & 0.430 & 0.800 \\
Top-5 img (Nem-CE$\oplus$Scout, Sonnet) & 0.432 & 0.535 & 0.533 & 0.527 & 0.600 \\
ReadAgent (Sonnet)                & 0.120 & 0.512 & 0.483 & 0.487 & 0.400 \\
\textbf{6-tool FC (Sonnet)}        & 0.682 & \textbf{0.644} & \textbf{0.551} & 0.591 & 0.800 \\
6-tool FC (Qwen3.5-27B)              & \textbf{0.745} & 0.633 & 0.527 & \textbf{0.611} & 0.800 \\
\midrule
$n_{\text{docs}}$              & 2 & 99 & 19 & 14 & 1 \\
\bottomrule
\end{tabular}
}
\caption{MMLongBench-Doc accuracy by document page-count bin (official scorer). \textbf{Bold} marks the best pipeline in each column; the $>$200 column is left unmarked because four pipelines tie there over a single document. Bins are uneven (most MMLB documents fall in 11--50 pages), so longest-bin estimates are noisy ($n_{\text{docs}}\in\{2,1\}$). The 6-tool FC agent stays flat with page count, while raw PDF drops only $-6.3$\,pp on the dense bins, far milder than the LDURL pattern because MMLB has fewer 100+ page documents.}
\label{tab:mmlb-page-count}
\end{table}

\section{Scout Reranker vs.\ Document Length}
\label{sec:appendix-scout}

The Nem-CE$\oplus$Scout cascade is the best retriever on MMLongBench but trails plain Nem-CE by $11.3$\,pp Hit@5 on LongDocURL (\S\ref{sec:exp-retrieval}). Table~\ref{tab:scout-length} shows this is a document-length effect. We bin LongDocURL by page count and compute Hit@5 on the answerable subset for both the plain Nem-CE retriever and the Nem-CE$\oplus$Scout cascade. Plain Nem-CE is essentially flat across lengths ($0.92$--$0.95$), but the cascade degrades with length, from a $-5.7$\,pp deficit on the shortest bin to $-22.0$\,pp on the longest. Plain Nem-CE's flatness shows the evidence page is still inside the $50$-page candidate set at every length, so the loss is in Scout's final selection: as the candidate set spans a longer document, Scout increasingly ranks the evidence page outside its top five. MMLongBench's documents are mostly short enough to avoid this regime, which is why the same cascade helps there ($+2.4$\,pp) and confirms the cross-benchmark asymmetry is a length artifact rather than a benchmark idiosyncrasy.

\begin{table}[t]
\centering
\scriptsize
\setlength{\tabcolsep}{5pt}
\begin{tabular}{@{}lrrrr@{}}
\toprule
Pages & $n$ & Nem-CE & Nem-CE$\oplus$Scout & $\Delta$ \\
\midrule
50--75    & 932 & 0.938 & 0.881 & $-5.7$  \\
75--100   & 555 & 0.944 & 0.813 & $-13.2$ \\
100--125  & 511 & 0.947 & 0.814 & $-13.3$ \\
125--150  & 287 & 0.916 & 0.697 & $-22.0$ \\
\midrule
Overall   & 2285 & 0.939 & 0.826 & $-11.3$ \\
\bottomrule
\end{tabular}
\caption{Hit@5 on the answerable LongDocURL subset by document page-count bin, for plain Nem-CE and the Nem-CE$\oplus$Scout cascade. Plain Nem-CE is flat with length; the cascade's recall deficit grows from $-5.7$ to $-22.0$\,pp, because Scout increasingly ranks the evidence page outside its final top five as the candidate set spans a longer document, even though the Nem-CE prefilter still contains it.}
\label{tab:scout-length}
\end{table}

\section{Effect of Thinking Mode}
\label{sec:appendix-thinking}

The main-paper results (Table~\ref{tab:headline}) disable the VLM's reasoning/thinking mode so that all configurations are compared under identical greedy decoding. Table~\ref{tab:thinking} isolates the effect of enabling a reasoning budget before the model answers, holding modality and retrieval depth fixed on MMLongBench-Doc. The sweep covers the two static image modes at three readers; we did not run the function-calling agent or Qwen3.5-27B with a reasoning budget, so no agent row appears here. This is why Table~\ref{tab:headline} holds every configuration at no-think: it is the only setting in which the agent and the static pipelines have matched decoding. Enabling thinking for the static modes alone would not be like-for-like, and the gap it opens is large enough to matter: Qwen3.5-9B all-images reaches $0.623$ with a reasoning budget, above the $0.616$ of the no-think Qwen3.5-27B agent, but the agent has no corresponding measurement.

\begin{table}[t]
\centering
\scriptsize
\setlength{\tabcolsep}{5pt}
\renewcommand{\arraystretch}{0.95}
\begin{tabular}{@{}llccc@{}}
\toprule
VLM & Input & No-think & Think & $\Delta$ \\
\midrule
\multirow{2}{*}{Qwen3.5-4B} & Top-5 img   & 0.512 & 0.553 & +4.1 \\
                            & All-images  & 0.526 & 0.578 & +5.2 \\
\midrule
\multirow{2}{*}{Qwen3.5-9B} & Top-5 img   & 0.541 & 0.561 & +2.0 \\
                            & All-images  & 0.559 & 0.623 & \textbf{+6.4} \\
\midrule
\multirow{2}{*}{Sonnet 4.5} & Top-5 img   & 0.502 & 0.543 & +4.1 \\
                            & All-images  & 0.477 & 0.514 & +3.7 \\
\bottomrule
\end{tabular}
\caption{Effect of enabling thinking mode (MMLongBench, AVG), with an 8k-token reasoning
budget. \textbf{Bold} marks the largest gain. Main-table numbers (Table~\ref{tab:headline}) use no-think throughout.}
\label{tab:thinking}
\end{table}

Thinking helps five of the six configurations we tested, by about $4$ to $6$\,pp; the sixth (Qwen3.5-9B on top-$k$ input, $+2.0$\,pp) has a paired interval that includes zero and we report it as a tie. The gains are largest where the input is largest: both open-weight readers gain more on all-images ($+5.2$\,pp at 4B, $+6.4$\,pp at 9B) than on top-$k$ input ($+4.1$ and $+2.0$\,pp), which fits the account that a whole-document input leaves more searching for a reasoning budget to do while a top-$k$ retriever has already localized the evidence. Sonnet gains about $4$\,pp on both inputs. No configuration gains more than about $6$\,pp, and the modality ordering is unchanged in every case, so reporting no-think numbers in the main table does not affect any of our conclusions.

\section{Retriever Cost and Architecture}
\label{sec:appendix-retriever-cost}

Table~\ref{tab:retriever-index-cost} reports the architectural details and cost profile for each retriever evaluated in Table~\ref{tab:retrieval-hit5}, including the index-encoder backbone and its size, embedding dimension, per-question LLM token cost at retrieval time, and per-document indexing cost; where a reranker is used it is named in the \textbf{Index encoder} column. Numbers are approximate single-document cost on one A100, and per-corpus totals scale linearly with $|D|$. BM25 indexes one document per page over the same MinerU markdown text used by the text-input rows, with lowercase word-regex tokenization and no stemming or stopword removal; the query is the question tokenized identically.

\begin{table*}[t]
\centering
\scriptsize
\setlength{\tabcolsep}{4pt}
\begin{tabular}{@{}llrlrrr@{}}
\toprule
Retriever & Index encoder & Enc.\ size & Embed dim & LLM tok/Q & Index time/doc & Index size/doc \\
\midrule
BM25 (MinerU)               & --- (lexical)        & ---     & sparse vocab    & 0          & ${<}1$\,s (CPU)  & ${<}1$\,MB \\
BGE-M3                      & XLM-RoBERTa-large    & 568M    & 1024 (dense)    & 0          & ${<}5$\,s (1 GPU) & ${\sim}3$\,MB \\
BM25 + Cohere 3.5          & --- + Cohere rerank  & ---     & sparse          & 0 (cross-enc.) & ${<}1$\,s        & ${<}1$\,MB \\
BGE-M3 + Cohere 3.5        & XLM-RoBERTa-l + Cohere & 568M  & 1024            & 0 (cross-enc.) & ${<}5$\,s        & ${\sim}3$\,MB \\
SIRA (full pipeline)        & BM25 + Claude Haiku~4.5 3-stage & --- & sparse        & ${\sim}11$k  & ${\sim}30$\,s (API) & ${\sim}5$\,MB \\
ColQwen2.5                  & Qwen2.5-VL-3B       & 3.0B    & 128 (late-int)  & 0          & ${\sim}30$\,s (1 GPU) & ${\sim}40$\,MB \\
Nem-CE                      & Nem-CE-3B           & 3.0B    & 128 (late-int)  & 0          & ${\sim}45$\,s (1 GPU) & ${\sim}40$\,MB \\
Nem-CE$\oplus$Scout & Nem-CE-3B + Qwen3.5-9B & 3.0B+9.0B & 128 (Nem-CE) & ${\sim}63$k & ${\sim}45$\,s (1 GPU) & ${\sim}40$\,MB \\
\bottomrule
\end{tabular}
\caption{Architecture and cost profile of each retriever in Table~\ref{tab:retrieval-hit5}. \textbf{Index encoder} names the backbone that produces document-side representations. \textbf{Enc.\ size} is the parameter count of that encoder. \textbf{Embed dim} is the per-token dimension for dense retrievers, ``sparse'' for lexical, and ``128 (late-int)'' for ColPali-family multi-vector encoders. \textbf{LLM tok/Q} is the LLM compute spent at retrieval time per question, on MMLongBench-Doc; for the Scout cascade it is dominated by the page images Scout reads (up to $50$ candidate pages at DPI~144, about $1{,}900$ image tokens each, a mean of $33$ pages per question), and it roughly doubles on LongDocURL, where nearly every document fills the $50$-page candidate set. \textbf{Index time/doc} and \textbf{Index size/doc} are one-time preprocessing for a whole PDF of tens to hundreds of pages, not per question: they are paid once per document and reused by every question on it, roughly eight on MMLongBench-Doc and six on LongDocURL, so ColQwen2.5's ${\sim}30$\,s is about $0.6$\,s per page at MMLongBench-Doc's 48-page mean. Pure-embedding retrievers (BM25, BGE-M3, ColQwen2.5, Nem-CE) incur zero LLM tokens at retrieval and only pay the precomputed-index cost. The Cohere 3.5 rows spend no generative-LLM tokens: the rerank is a cross-encoder call over the question and the top-$50$ candidate pages' text, so its cost is a fixed cross-encoder pass rather than LLM generation, and it matches the multi-stage SIRA pipeline ($\sim$11k LLM tokens per question) within 0.3\,pp on MMLongBench. Image-based retrievers lead text retrievers by about 17\,pp between base encoders and by about 9\,pp best-to-best, and the best image retriever (plain Nem-CE) does so with no LLM in the loop; the Scout cascade adds a large per-question LLM cost for a net loss against it (Table~\ref{tab:retrieval-hit5}).}
\label{tab:retriever-index-cost}
\end{table*}

\section{Retrieval-Depth Sweep}
\label{sec:appendix-ksweep}

We sweep the number of retrieved pages $k$ for the ColQwen2.5 image retriever across Sonnet and all three Qwen3.5 sizes on both benchmarks (Table~\ref{tab:ksweep}), reporting accuracy and mean input tokens; \textbf{Full} sends the whole document. Accuracy rises steeply to about $k{=}10$ and is flat thereafter: Sonnet peaks at $k{=}10$ and declines by $k{=}20$ as extra pages add distractors, while the Qwen readers gain at most $0.8$\,pp between $k{=}10$ and $k{=}20$ for $2\times$ the input tokens. Our default of $k{=}5$ sits just below the peak and captures most of its accuracy at roughly half the tokens, which is why we use it throughout. Sending more than about ten pages is not worth the added cost, and even Full (the whole document) never beats the best $k$, though for the local Qwen readers on MMLongBench it comes within a point of it. The LDURL Full cell for Qwen3.5-27B is omitted, as is the same configuration's all-images cell in Table~\ref{tab:headline}: rendering the uncapped page images of the longest documents did not complete, and an image-size-capped rerun would not be token-comparable to the other cells. This sweep also provides a token-matched control for the agent-versus-static comparison: the $k{=}20$ cells give the static pipeline \emph{more} input tokens than the corresponding agent consumes, and the agent still leads at Sonnet ($+11.6$\,pp, significant) though only by $+2.1$\,pp at Qwen3.5-27B, where the interval includes zero (paired tests in \S\ref{sec:appendix-significance}), so the agent's advantage is the access policy rather than the token budget.

\begin{table*}[t]
\centering
\scriptsize
\setlength{\tabcolsep}{4pt}
\begin{tabular}{@{}ll ccccc@{}}
\toprule
Benchmark & Reader & $k{=}2$ & $k{=}5$ & $k{=}10$ & $k{=}20$ & Full \\
\midrule
\multirow{4}{*}{MMLB}
 & Sonnet      & 0.469 / 3.1k & 0.502 / 7.6k & \textbf{0.514} / 15k & 0.508 / 27k & 0.477 / 32k \\
 & Qwen3.5-4B  & 0.484 / 3.6k & 0.512 / 9.0k & 0.528 / 18k & \textbf{0.530} / 35k & 0.526 / 33k \\
 & Qwen3.5-9B  & 0.500 / 3.6k & 0.541 / 9.0k & 0.560 / 18k & \textbf{0.564} / 35k & 0.559 / 33k \\
 & Qwen3.5-27B & 0.523 / 3.6k & 0.565 / 9.0k & 0.594 / 18k & \textbf{0.595} / 34k & 0.588 / 30k \\
\midrule
\multirow{4}{*}{LDURL}
 & Sonnet      & 0.608 / 3.0k & 0.641 / 7.5k & \textbf{0.644} / 15k & 0.628 / 29k & 0.360 / 44k \\
 & Qwen3.5-4B  & 0.572 / 3.9k & 0.590 / 9.6k & 0.598 / 19k & \textbf{0.606} / 38k & 0.569 / 66k \\
 & Qwen3.5-9B  & 0.585 / 3.9k & 0.605 / 9.6k & \textbf{0.610} / 19k & 0.609 / 38k & 0.592 / 66k \\
 & Qwen3.5-27B & 0.589 / 3.9k & 0.615 / 9.4k & 0.623 / 19k & \textbf{0.623} / 37k & --- \\
\bottomrule
\end{tabular}
\caption{Retrieval-depth sweep with ColQwen2.5 (accuracy / mean input tokens). \textbf{Bold} marks each row's best $k$. MMLB uses the extractor scorer, LDURL the official scorer (Table-1 scale). Accuracy peaks near $k{=}10$ and is flat-to-down by $k{=}20$ while tokens keep growing, so $k{=}5$ is a favourable accuracy/cost point. This is a self-contained sweep with the retriever fixed to ColQwen2.5, so cells differ from Table~\ref{tab:headline}, whose per-reader top-5 rows use the best retriever for each reader and whose values may differ by up to $\sim$1\,pp of decoding variance from an independent run.}
\label{tab:ksweep}
\end{table*}

\section{Agent Implementation Details}
\label{sec:appendix-agents}

This appendix expands the description of the two agentic configurations evaluated in \S\ref{sec:exp-agents}: our six-tool function-calling agent and the ReadAgent baseline.

\subsection{Six-Tool Function-Calling Agent}
\label{sec:appendix-fc}

The function-calling agent wraps the same VLM used for static configurations in a multi-turn loop. The VLM is given a fixed toolkit and a per-question budget. At each turn it either issues one or more tool calls or commits a final answer via the \texttt{finish} tool.

\paragraph{Toolkit.}
The agent has access to six tools:
\begin{itemize}
\item \textbf{\texttt{get\_catalog}} returns a deterministic structural map of the document built from MinerU's per-page layout output, with a per-page index of text size and figure/table counts, a flat section list with title and page range, and flat figure and table lists with captions, inferred neighbor text, and stable integer ids. No LLM is involved in the catalog construction, and the same MinerU output always produces the same catalog.
\item \textbf{\texttt{read\_pages(pages, resolution)}} renders the requested pages as image content blocks at a resolution the model chooses per call: \texttt{medium} (longest side $700$\,px) or \texttt{full} (longest side $1{,}400$\,px, rendered at DPI 180). Sonnet 4.5 asks for \texttt{full} on $62\%$ of its page reads, Qwen3.5-4B on $68\%$, and Qwen3.5-27B on $12\%$. Multiple pages can be requested in a single call.
\item \textbf{\texttt{get\_figure(figure\_id)}} returns the MinerU-extracted figure crop at the given catalog id as an image, plus its caption and neighbor text. Falls back to the full page image when the crop is unavailable.
\item \textbf{\texttt{get\_table(table\_id)}} returns the MinerU-extracted table crop, its caption and neighbor text, and the table's structured HTML. The HTML preserves cell boundaries and is more robust than image-only access for numeric reasoning.
\item \textbf{\texttt{search\_text(query, top\_k)}} runs BM25 over MinerU per-page text and returns the top-$k$ page numbers, each with a matching text excerpt. Used by the VLM as an in-loop retrieval action when it has formed a more specific query than the original question.
\item \textbf{\texttt{finish(answer)}} commits the final answer and ends the loop.
\end{itemize}

The catalog is also seeded into the very first user turn alongside the question, so the VLM has the document map without paying a tool-call round-trip, and explicit \texttt{get\_catalog} calls are available for re-fetching. Seeding is therefore a latency optimisation rather than a source of the accuracy gain: the same information is reachable by tool call either way, and the agent would simply spend one extra turn retrieving it.

\paragraph{Loop structure.}
At each turn the VLM sees the conversation so far (including all prior tool returns) and chooses one or more tool calls. We cap the agent at eight tool calls per question. The budget is checked between turns, so a turn that issues several calls in parallel can overshoot it slightly. The Sonnet runner ends the loop at the first \texttt{finish}. For the Qwen3.5 runs the budget is stated in the prompt, the loop is bounded by a hard limit of $20$ model calls, and it ends when the model returns a turn that issues no tool call: \texttt{finish} records the answer but is itself a tool, so a reader that keeps calling it keeps the loop alive until that ceiling. Qwen3.5-4B exits on its own on $64\%$ of questions and exhausts the $20$-call ceiling on the other $36\%$; of those, $97\%$ had already emitted \texttt{finish} with an answer. Its call distribution is bimodal, with a median of $5$ and a spike of $390$ questions at exactly $20$. Qwen3.5-9B and 27B reach the ceiling on $8\%$ of questions. This is the cost asymmetry reported in \S\ref{sec:exp-agents}. If the budget is exhausted without a \texttt{finish}, the VLM is given a final turn that must call \texttt{finish}. The temperature is set to zero throughout, and we use each provider's native function-calling API (the Claude API for Sonnet, and vLLM's OpenAI-compatible \texttt{/v1/chat/completions} endpoint with tool schemas for Qwen3.5).

\paragraph{Cost profile.}
On MMLongBench-Doc with Claude Sonnet 4.5, the agent averages 2.68 tool calls per question, 20.1\,k input tokens (catalog plus tool returns), and 538 output tokens (mostly tool argument JSON plus the final answer). The eight-call budget is rarely exhausted, and the modal trajectory uses two tool calls. We set the cap at eight to leave headroom rather than to constrain the policy, and the observed distribution confirms it does not bind: a cap of six would truncate $3.6\%$ of trajectories, seven $0.8\%$, and eight $0.1\%$, while the longest trajectory we observe uses nine calls. Eight therefore sits just past the point where a lower cap would begin to change answers, so the results are insensitive to the exact value.

\subsection{ReadAgent Baseline}
\label{sec:appendix-readagent}

ReadAgent~\cite{lee2024readagent} is a gist-based agent design that we re-implement as the closest published baseline to our toolkit. It pre-computes per-page ``gist memories'' and lets the VLM iteratively expand the most relevant gists.

\paragraph{Pipeline.}
For each document, ReadAgent first generates a short gist for every page using a small LLM (Claude Haiku~4.5 in our setup), with a fixed prompt and a 60-word target. At question time, the VLM is shown the full sequence of gists with the question and asked to select the most relevant pages to read in full. The selected pages are rendered at DPI=144 and concatenated with the gists for a final answer turn.

\paragraph{Hyperparameters.}
Gist length is capped at 60 words, lookup chooses 1--7 pages (median 2 in our runs), and both gisting and lookup run at temperature zero. Gists are cached on disk per-document and reused across all questions for that document, so the gisting cost amortizes over the full benchmark.

\paragraph{Key design difference vs.\ our toolkit.}
ReadAgent's lookup operates on LLM-generated gists, which compress each page into 60 words of free-form text, and once pages are selected, no further tool calls happen. Our six-tool agent instead seeds the VLM with a deterministic structural catalog (MinerU-derived section, figure, and table indices) and exposes structured figure-fetch, table-fetch, and BM25 search actions, so the VLM can explore specific structured content rather than committing to a page set up front from compressed text summaries. This difference is reflected in the agentic accuracy gap (\S\ref{sec:exp-agents}). ReadAgent's gist memory loses fine-grained list and table information, while the structural catalog plus targeted tools preserves it.

\paragraph{Reproducibility.}
Both agents run with deterministic decoding ($T=0$). Tool returns are byte-deterministic given the same MinerU output and PDF rendering, and on a fixed VLM checkpoint repeated runs of the same question yield the same trajectory in our spot checks. The MinerU outputs, page renderings, and gist caches are produced once per document and reused across all VLM configurations, so reported per-question latencies reflect agent-loop overhead and not preprocessing time.

\section{Architectural Cost of the Comparison Agents}
\label{sec:appendix-agent-cost}

Table~\ref{tab:agent-cost} places the agentic systems of Table~\ref{tab:headline} on both axes at once, MMLongBench accuracy alongside what each one costs to run. Read together they make the practitioner-facing point of \S\ref{sec:exp-agents}: our agent carries the least machinery of any system in the group, one VLM, a median of two tool calls, and a preprocessing step with no LLM in it, and is nonetheless level with or ahead of every system scored under the official scorer.

Two contrasts carry the cost argument. SimpleDoc keeps two large models in the loop per question, in its published configuration a $30$B reasoner driving a $32$B answerer over several retrieve-and-refine rounds, where we use a single VLM at $20$\,k input tokens. ReadAgent, the closest gist-based design, spends roughly one Claude Haiku~4.5 call \emph{per page} to build its gist memory (\S\ref{sec:appendix-readagent}), $5{,}775$ calls over the $135$ MMLongBench-Doc documents, a mean of $43$ per document, or ${\sim}5.3$ extra LLM calls per question once amortized over that document's questions. Our catalog is derived from a single MinerU layout parse with no LLM in the loop.

Rows are ordered by accuracy, which is also how the protocol caveat becomes visible: the top row is DocLens with Gemini-2.5-Pro, $5.1$\,pp above our Sonnet agent under DocLens's own protocol, a gap smaller than the $8.9$\,pp that re-scoring SimpleDoc under the official scorer removed (\S\ref{sec:exp-agents}), so the two cannot be ordered across protocols.

\begin{table*}[t]
\centering
\scriptsize
\setlength{\tabcolsep}{4pt}
\renewcommand{\arraystretch}{1.05}
\begin{tabularx}{\textwidth}{@{}p{2.7cm}lcXXX@{}}
\toprule
System & Reader & MMLB & Models in the loop & Per-question work & Offline preprocessing \\
\midrule
DocLens$^\S$~\cite{zhu2025doclens}   & Gemini-2.5-Pro & 0.676 & 3 agents (navigator, localizer, reasoner) & one pass per role & page index \\
DocLens$^\S$~\cite{zhu2025doclens}   & Claude-4-Sonnet & 0.633 & 3 agents (navigator, localizer, reasoner) & one pass per role & page index \\
\textbf{6-tool FC (ours)} & \textbf{Sonnet 4.5} & \textbf{0.625} & \textbf{1 VLM} & \textbf{median 2 tool calls, 20\,k tok} & \textbf{MinerU catalog, no LLM} \\
\textbf{6-tool FC (ours)} & \textbf{Qwen3.5-27B} & \textbf{0.616} & \textbf{1 VLM} & \textbf{median 2 tool calls, 33\,k tok} & \textbf{MinerU catalog, no LLM} \\
SimpleDoc~\cite{jain2025simpledoc} (our re-run) & Sonnet 4.5 & 0.608 & 2 (both Sonnet 4.5) & iterative retrieve-and-refine rounds & dual-cue page index \\
ReadAgent~\cite{lee2024readagent} & Sonnet 4.5 & 0.498 & 1 VLM + a gisting LLM & lookup turn plus answer turn & 1 LLM call \emph{per page} (43/doc) \\
MDocAgent$^\dagger$~\cite{han2025mdocagent} & Qwen2-VL-7B & 0.315 & 5 specialized agents + retriever & one pass per agent, then aggregation & ColPali page index \\
\bottomrule
\end{tabularx}
\caption{Agentic systems of Table~\ref{tab:headline} by MMLongBench accuracy and architectural cost, ordered by accuracy. \textbf{Bold} marks our own rows, not the column best. $^\dagger$/$^\S$ mark scores under the system's own protocol rather than the official scorer ($\dagger$ GPT-4o judge, $\S$ DocLens's own); re-scoring SimpleDoc under the official scorer costs $8.9$\,pp (\S\ref{sec:exp-agents}); unmarked rows use the official scorer. Our rows and the ReadAgent row are measured from our own runs; the DocLens and MDocAgent entries, and SimpleDoc's published 30B/32B configuration, are as reported in those papers, since SimpleDoc is the only external system we re-ran. The SimpleDoc row here is that re-run, with Sonnet 4.5 in both roles.}
\label{tab:agent-cost}
\end{table*}

\section{Tool Usage Analysis for the 6-Tool FC Agent}
\label{sec:appendix-tool-usage}

To understand which tools contribute most to the 6-tool function-calling agent's accuracy, we analyze the per-question tool-call traces from the Sonnet 4.5 run on MMLongBench-Doc (N=1{,}082, mean 2.68 tool calls per question). Table~\ref{tab:tool-usage-per-evidence} reports how often each tool is invoked, broken down by evidence type.

\begin{table}[t]
\centering
\scriptsize
\setlength{\tabcolsep}{4pt}
\begin{tabular}{@{}lrrrrr@{}}
\toprule
Evidence type & N & search & read & figure & table \\
\midrule
TXT (Pure-text)    & 305 & 66\% & 86\% & 13\% & 14\% \\
LAY (Layout)       & 119 & 60\% & 92\% & 10\% & 8\%  \\
CHA (Chart)        & 178 & 69\% & 79\% & \textbf{34\%} & 8\% \\
TAB (Table)        & 218 & 72\% & 61\% & 5\%  & \textbf{51\%} \\
FIG (Figure)       & 304 & 52\% & 81\% & 25\% & 8\%  \\
\bottomrule
\end{tabular}
\caption{Per-evidence-type tool invocation rates in the Sonnet 4.5 FC agent on MMLongBench-Doc. Values are the percentage of questions for which the agent invoked each tool at least once. The agent reliably routes table-evidence questions to \texttt{get\_table} (51\%) and chart/figure questions to \texttt{get\_figure} (34\%/25\%), evidence that the structured catalog enables targeted access rather than uniform page reading.}
\label{tab:tool-usage-per-evidence}
\end{table}

\paragraph{Tool-call budget.}
Across 1,082 questions, the median trajectory uses 2 tool calls and the mean is 2.68, with 76\% of questions finishing within 3 tool calls and 86\% within 4. The full distribution over the number of (non-finish) tool calls is 0 (n=15), 1 (240), 2 (377), 3 (187), 4 (113), 5 (57), 6 (54), 7 (30), 8 (8), 9 (1). The budget is eight tool calls, checked between turns, so a turn that requests several pages or figures in parallel can push the per-question count slightly past 8. Only 3.6\% of questions issue more than 6 tool calls, so a tighter 6-call budget would affect very few trajectories and is a defensible cost-saving operating point if a deployment is latency- or cost-sensitive.

\section{Statistical Significance}
\label{sec:appendix-significance}

To quantify the uncertainty on Table~\ref{tab:headline} we bootstrap the per-question scores: for each configuration we resample questions with replacement $10{,}000$ times, under a fixed seed, and take the $2.5$th and $97.5$th percentiles of the resampled mean. The resulting $95\%$ confidence half-widths are uniform within each benchmark, $\pm 0.028$--$0.030$ on MMLongBench-Doc ($N{=}1{,}082$) and $\pm 0.017$--$0.019$ on LongDocURL ($N{=}2{,}325$). A useful rule of thumb follows: on these benchmarks a difference below ${\sim}3$\,pp (MMLB) or ${\sim}2$\,pp (LDURL) is within single-configuration noise.

We test the paper's central claims with a paired bootstrap over the questions both configurations answer, resampling the per-question score differences and calling a gap significant when its $95\%$ interval excludes zero (Table~\ref{tab:significance}). On MMLongBench the headline results are significant: holding the reader fixed, the 6-tool Sonnet agent beats Top-5 ColQwen with the same reader by $+11.7$\,pp, and it also beats the strongest \emph{retrieval} row at any reader (Top-5 Nem-CE$\oplus$Scout with Qwen3.5-9B) by $+6.1$\,pp, so the lead survives comparison against a stronger reader than its own and not only against a same-reader one. Against the strongest static row overall, all-images with Qwen3.5-27B ($0.588$), the agent leads by $+3.7$\,pp, $[+0.6,+6.8]$. It beats ReadAgent ($+12.6$\,pp), and the image-over-text retrieval gap (ColQwen over BM25) holds on both benchmarks ($+10.0$ / $+8.7$\,pp). On LongDocURL the picture is different: the agent's $+1.5$\,pp margin over the best static pipeline, Top-5 Nem-CE, has an interval that includes zero, so on that benchmark the two are level and we do not order them. Three further gaps are \emph{not} significant and we likewise report them as ties: all-images versus MinerU-text on MMLongBench ($-0.4$\,pp, $p{=}0.80$), Sonnet versus Qwen3.5-27B as the agent reader ($+0.9$\,pp), and top-$k$ retrieval versus all-images at both open-weight readers ($+2.5$ and $+0.5$\,pp). The Scout reranker's effect is significant in \emph{both} directions: it helps on MMLongBench ($+2.5$\,pp) and hurts on LongDocURL ($-2.7$\,pp), confirming the cross-benchmark asymmetry is real and not noise. Likewise the SIRA-versus-single-rerank comparison is a zero-centered tie ($+0.1$\,pp Hit@5, $[-2.8,+2.9]$ against BM25+Cohere; $-0.4$\,pp $[-3.3,+2.6]$ against BGE-M3+Cohere), so the two are statistically indistinguishable at our $2$\,pp noise floor and we describe the single rerank as equivalent rather than merely close.

Three of the table's rows address potential confounds directly. \emph{Token matching:} the agent's win is not a budget effect. Static top-$k$ accuracy peaks near $k{=}10$ and declines at $k{=}20$ as added pages act as distractors (Table~\ref{tab:ksweep}), so a static pipeline given the agent's token budget or more still loses: the Sonnet agent ($0.625$ at $20$k input tokens) beats static ColQwen $k{=}20$ ($0.508$ at $27$k) by $+11.6$\,pp paired, while the Qwen3.5-27B agent ($0.616$ at $33$k) is only $+2.1$\,pp above its static $k{=}20$ ($0.595$ at $34$k), an interval that includes zero. At Sonnet, then, the advantage is the access policy rather than the compute; at Qwen3.5-27B the two are indistinguishable once tokens are matched. \emph{Pure-text evidence:} the image-retriever advantage is not only about tables and figures. Restricted to questions whose gold evidence carries no figure, chart, or table (that is, plain-text and layout-bearing text pages only), Nem-CE still beats the best text pipeline (BGE-M3+Cohere) on Hit@5, $0.889$ vs.\ $0.833$ on MMLongBench and $0.925$ vs.\ $0.885$ on LongDocURL, so page renderings preserve signal that text extraction loses even for textual evidence, though the margin is smaller than on visual evidence. \emph{Extractor bias:} scoring raw responses instead of extracted answers under the official scorer costs the static pipelines and ReadAgent $47.1$--$49.7$\,pp (their long-form prose fails string matching, scoring near zero raw) but costs the FC agent only $0.7$\,pp ($0.631$ vs.\ $0.625$), since its \texttt{finish} tool already returns a short structured answer. Table~\ref{tab:extractor-bias} gives the per-system breakdown. The extractor is what makes the baselines' numbers comparable at all, while for the agent it is a small penalty, so it cannot be inflating the agent's lead; if anything it understates it.

\begin{table}[t]
\centering
\scriptsize
\setlength{\tabcolsep}{4pt}
\begin{tabular}{@{}lccc@{}}
\toprule
System (Sonnet 4.5) & Raw & Extracted & Extractor gain \\
\midrule
6-tool FC agent & 0.631 & 0.625 & $\mathbf{-0.7}$\,pp \\
All-images       & 0.005 & 0.477 & $+47.3$\,pp \\
Top-5 ColQwen    & 0.006 & 0.502 & $+49.6$\,pp \\
MinerU text      & 0.009 & 0.480 & $+47.1$\,pp \\
ReadAgent        & 0.002 & 0.498 & $+49.7$\,pp \\
\bottomrule
\end{tabular}
\caption{Effect of the shared answer extractor on MMLongBench-Doc ($n{=}1{,}082$), scoring each system's raw response and its extracted short answer under the same official scorer. Extraction is part of the benchmark's own published protocol, and we apply its released prompt unchanged to every system; the near-zero raw column reflects the scorer's whole-string comparison rather than wrong answers, e.g.\ for gold ``Less well-off'' a response containing ``\ldots their children will be \emph{less well off} financially\ldots'' scores $0.0$ raw and $0.92$ extracted. Four unrelated baselines are lifted by nearly the same amount because their raw free-form prose scores near zero under string matching; the agent is the only system the extractor does not help, since its \texttt{finish} tool already emits a short structured answer. The protocol therefore favours the baselines and understates the agent's margin. The extracted column reproduces the corresponding Table~\ref{tab:headline} cells for all-images, MinerU text, and ReadAgent; the ColQwen row is the $k{=}5$ depth-sweep run of Table~\ref{tab:ksweep}, so its gain is computed within that run. The agent's \texttt{finish} tool already emits a short answer, so its own output stands in for the raw column.}
\label{tab:extractor-bias}
\end{table}

\begin{table}[t]
\centering
\scriptsize
\setlength{\tabcolsep}{4pt}
\begin{tabularx}{\columnwidth}{@{}Xrl@{}}
\toprule
Comparison & $\Delta$ (pp) & 95\% CI \\
\midrule
Agent $>$ Top-5 ColQwen (same reader), MMLB & $+11.7$ & $[+8.8,+14.8]$ \\
Agent $>$ best Top-5 row, any reader, MMLB$^{b}$ & $+6.1$ & $[+3.1,+9.2]$ \\
Agent $>$ ReadAgent, MMLB               & $+12.6$ & $[+9.6,+15.7]$ \\
Agent vs best static (Top-5 Nem-CE), LDURL (n.s.) & $+1.5$  & $[-0.3,+3.1]$  \\
Agent-27B $>$ Top-5 9B, MMLB            & $+5.2$  & $[+2.2,+8.2]$ \\
Raw PDF $>$ all-images, MMLB            & $+4.5$  & $[+1.7,+7.3]$  \\
ColQwen $>$ BM25, MMLB                  & $+10.0$ & $[+7.5,+12.6]$ \\
ColQwen $>$ BM25, LDURL                 & $+8.7$  & $[+7.2,+10.2]$ \\
Scout helps Nem-CE, MMLB                & $+2.5$  & $[+0.4,+4.6]$  \\
Scout hurts Nem-CE, LDURL               & $-2.7$  & $[-4.1,-1.5]$  \\
Agent $>$ static @ matched tokens (Sonnet), MMLB & $+11.6$ & $[+8.7,+14.6]$ \\
Agent-27B vs static @ matched tokens (n.s.) & $+2.1$  & $[-0.9,+5.2]$  \\
Image $>$ text retr., pure-text evid., MMLB$^{h}$ & $+5.6$ & $[+0.6,+11.1]$ \\
Image $>$ text retr., pure-text evid., LDURL$^{h}$ & $+4.0$ & $[+2.3,+5.7]$ \\
All-images vs MinerU-text, MMLB (n.s.)  & $-0.4$  & $[-3.5,+2.6]$  \\
Agent Sonnet vs 27B, MMLB (n.s.)        & $+0.9$  & $[-1.6,+3.4]$  \\
SIRA vs BM25+Cohere rerank, MMLB$^{h}$ (equiv.) & $+0.1$ & $[-2.8,+2.9]$ \\
\bottomrule
\end{tabularx}
\caption{Paired-bootstrap significance tests ($10{,}000$ resamples of per-question score differences). ``Agent'' is our 6-tool FC agent with Sonnet 4.5 and ``Agent-27B'' the same toolkit with Qwen3.5-27B; ReadAgent is named in full where it appears. A comparison is significant when its $95\%$ interval excludes zero; ``n.s.'' marks gaps within noise that we report as ties, and ``equiv.'' a zero-centered tie we report as statistical equivalence. The first row holds the reader fixed (Sonnet 4.5 on both sides); $^{b}$compares the Sonnet agent against the strongest retrieval row of Table~\ref{tab:headline} at any reader, Top-5 Nem-CE$\oplus$Scout with Qwen3.5-9B ($0.564$), rather than the larger same-reader gap. The strongest static row overall is all-images with Qwen3.5-27B ($0.588$), which the Sonnet agent leads by $+3.7$\,pp paired, $[+0.6,+6.8]$. ``Matched tokens'' compares the agent against static ColQwen top-$20$, which consumes \emph{more} input tokens than the agent ($27$k vs.\ $20$k for Sonnet; $34$k vs.\ $33$k for Qwen3.5-27B). $^{h}$Hit@5 comparisons: the pure-text-evidence rows compare Nem-CE against BGE-M3+Cohere on questions whose gold evidence carries no figure, chart, or table, i.e.\ plain-text and layout-bearing text pages ($n{=}180$ MMLB, $n{=}1{,}069$ LDURL); the SIRA row compares its full multi-stage pipeline against a single Cohere rerank over the same $854$ answerable MMLB questions.}
\label{tab:significance}
\end{table}

\section{Latency}
\label{sec:appendix-latency}

Table~\ref{tab:headline} reports input tokens as the cost axis; Table~\ref{tab:latency} adds measured per-question wall-clock latency from the logged inference time of each run. The ordering mirrors the token cost but is not identical. Top-$k$ retrieval is the cheapest mode and among the fastest (median $5.2$\,s with Nem-CE and $6.4$\,s with ColQwen2.5, against $5.1$\,s for ReadAgent's two-turn lookup), and only image-free MinerU text is faster still (median $3.5$\,s), which sends four times the tokens but no images and so skips image decoding entirely; the 6-tool agent is competitive with all-images despite issuing several sequential tool calls (median about $11$\,s), and raw PDF is the slowest (median $12.9$\,s on MMLB, $20.5$\,s on LDURL) because the API ingests every page as both an image and a text layer. Retrieval's speed advantage compounds its token advantage: it is roughly $2\times$ faster than the agent and $2$--$4\times$ faster than raw PDF, at about a third of the agent's input tokens and a tenth or less of raw PDF's.

These latencies are inference-time only. The one-time preprocessing that retrieval and the agent depend on (MinerU parse, page rendering, index/embedding build, catalog construction) is computed once per document and amortized across every question on that document, which on these benchmarks averages ${\sim}8$ questions per document on MMLongBench ($1{,}082/135$) and ${\sim}5.9$ on LongDocURL ($2{,}325/396$). Per-retriever index-build cost is detailed in Table~\ref{tab:retriever-index-cost}: BM25 indexes in under a second on CPU, and the image retrievers embed a document once on a single A100. The heaviest preprocessing in the table is not ours but the ReadAgent baseline's, which spends roughly one Claude Haiku~4.5 call \emph{per page} to build its gist memory (\S\ref{sec:appendix-readagent}): $5{,}775$ gisting calls across the $135$ MMLongBench-Doc documents, a mean of $43$ per document, or ${\sim}5.3$ extra LLM calls per question once amortized, roughly $2.7\times$ its own per-question inference. Our catalog, by contrast, is built from MinerU layout output with no LLM in the loop. Counting preprocessing in would therefore widen our margin over that baseline rather than narrow it. The three access modes differ in what they require offline, which the per-question token column does not show. Raw PDF and all-images need nothing. Top-$k$ retrieval needs a per-document embedding pass, ${\sim}30$\,s for ColQwen2.5 and ${\sim}45$\,s for Nem-CE on one A100 at ${\sim}40$\,MB per document, and the LLM-based rerank pipelines (SIRA, Nem-CE$\oplus$Scout) additionally spend generative tokens at query time, while the Cohere cross-encoder rerank does not. Our agent needs only the MinerU parse that the text-input modes already require, with no embedding index and no LLM. Amortized over the ${\sim}8$ and ${\sim}5.9$ questions per document these benchmarks provide, none of this reverses the token- or latency-cost ordering, but it does mean the token column understates retrieval's full cost relative to the agent's.

\begin{table}[t]
\centering
\scriptsize
\setlength{\tabcolsep}{4pt}
\begin{tabular}{@{}lrrr@{}}
\toprule
Configuration (Sonnet 4.5) & Median (s) & Mean (s) & Mean tok \\
\midrule
Top-5 ColQwen, MMLB       & 6.4  & 7.0  & 7.6k  \\
Top-5 Nem-CE, MMLB        & 5.2  & 5.6  & 7.6k  \\
MinerU text, MMLB         & 3.5  & 3.8  & 27k   \\
ReadAgent, MMLB           & 5.1  & 5.4  & 11k   \\
6-tool FC agent, MMLB     & 11.4 & 13.7 & 20k   \\
All-images, MMLB          & 10.1 & 11.5 & 32k   \\
Raw PDF, MMLB             & 12.9 & 15.8 & 80k   \\
\midrule
Top-5 ColQwen, LDURL      & 4.9  & 5.4  & 7.5k  \\
6-tool FC agent, LDURL    & 10.6 & 12.8 & 21k   \\
All-images, LDURL         & 10.6 & 13.4 & 44k   \\
Raw PDF, LDURL            & 20.5 & 22.6 & 150k  \\
\bottomrule
\end{tabular}
\caption{Measured per-question wall-clock latency (from logged inference time) alongside mean input tokens. Retrieval is the cheapest mode on both benchmarks and among the fastest; on MMLongBench, where we timed every mode, only image-free MinerU text is faster, with ReadAgent's two-turn lookup in the same range as retrieval. The agent is competitive with all-images and faster than raw PDF on both. Latencies are inference-time only; one-time preprocessing is amortized over all questions on a document (\S\ref{sec:appendix-latency}).}
\label{tab:latency}
\end{table}

\section{Per-Question Routing Analysis}
\label{sec:appendix-router}

\paragraph{Oracle definition and membership.}
Given pipelines $\{S_1,\dots,S_m\}$ scored on the same questions, the per-question
oracle's score on $q$ is $\max_j \text{score}(S_j, q)$. Averaged over the benchmark
this is the accuracy a perfect selector would achieve if it knew, with hindsight,
which pipeline to trust per question. Unless stated otherwise the member set is three
of our strongest configurations: the 6-tool FC agent with Sonnet 4.5, the same agent
with Qwen3.5-27B, and Top-5 Nem-CE retrieval with Sonnet. Because it is a
per-question maximum the oracle is computed only over questions all members answer,
and it rises mechanically with the number of members, so we hold the member set
fixed at three throughout rather than reporting a figure that grows with how many
pipelines we happen to have run.

This oracle gains roughly $13$\,pp over the best single pipeline, so the pipelines succeed on different questions. Table~\ref{tab:headline}'s oracle row, computed over its own common subset, implies $+13.1$\,pp on MMLongBench and $+12.9$\,pp on LongDocURL against the best single pipeline there. We test whether a router conditioned on evidence type can capture this headroom. For every evidence type we identify the pipeline with the highest accuracy on that type, then route each question to its type's best pipeline; using the \emph{gold} evidence label makes this an upper bound on any type-conditioned router. Unlike Table~\ref{tab:per-evidence}, which tags a question with every evidence type its gold pages carry, this analysis assigns each question to a single mutually-exclusive bucket so that a router has exactly one branch to take per question, which shifts the per-type means slightly. The 6-tool agent is the strongest pipeline within every evidence-type stratum on MMLongBench except charts, where ColQwen edges it by $1.5$\,pp (e.g.\ it leads on tables at $0.703$ over MinerU-text at $0.616$, and on figures at $0.529$ over ColQwen at $0.465$), and on LongDocURL for every type except tables, where Top-5 Nem-CE leads by $1.6$\,pp, and a $3$-question ``other'' bucket; the type-ceiling router therefore assigns the agent almost everywhere and gains only $+0.2$\,pp on MMLongBench and $+0.6$\,pp on LongDocURL over simply running the agent, both within noise. A realistic router that conditions on observable features (question keywords and answer format, without the gold label) is \emph{worse} than the agent alone ($-0.9$\,pp on MMLongBench, $-7.1$\,pp on LongDocURL), since it sometimes routes away from the agent. The oracle headroom is therefore real but \emph{intra}-type: it lives in disagreements among the pipelines on individual questions of the same evidence type, not in any type-to-pipeline mapping. Recovering it requires a learned per-question router with a finer signal than evidence type, which we leave to future work; a fixed type-based routing rule does not beat a single strong agent. Table~\ref{tab:routing} summarizes the four strategies.

\begin{table}[t]
\centering
\scriptsize
\setlength{\tabcolsep}{4pt}
\begin{tabular}{@{}lccc@{}}
\toprule
Strategy & MMLB & LDURL & vs.\ best single \\
\midrule
Best single pipeline (6-tool FC)   & 0.627 & 0.661 & reference \\
Gold-type router (upper bound)     & 0.629 & 0.667 & $+0.2$ / $+0.6$\,pp \\
Observable-feature router          & 0.618 & 0.590 & $-0.9$ / $-7.1$\,pp \\
Per-question oracle (hindsight)    & 0.752 & 0.796 & $+12.5$ / $+13.5$\,pp \\
\bottomrule
\end{tabular}
\caption{Routing strategies, all with Sonnet 4.5, scored on the questions all four candidate pipelines answer ($n{=}1{,}064$ MMLB, $n{=}2{,}287$ LDURL). The gold-type router uses the gold evidence label and is therefore an upper bound on any type-conditioned rule; it gains under a point, because it routes to the agent almost everywhere. The observable-feature router uses only question keywords and answer format. The agent reads $0.627$/$0.661$ here because these rows use the four-pipeline shared subset rather than the full question set of Table~\ref{tab:headline} ($0.625$/$0.659$). This oracle is over these four pipelines rather than the three configurations of Table~\ref{tab:headline}'s oracle row; the roughly $13$\,pp headroom conclusion is unchanged.}
\label{tab:routing}
\end{table}

\section{Chunked Map-Reduce for Long Documents}
\label{sec:appendix-mapreduce}

The all-images mode collapses on long documents (\S\ref{sec:appendix-page-count}) because the API stitches and downsamples pages once a request exceeds its $100$-image/size cap. We test whether this is an artifact of the single-request implementation rather than a limit of the all-images access mode itself, by splitting each long document into contiguous chunks of at most $100$ pages, answering the question over each chunk independently (rendered at the same $144$\,DPI, $1024$\,px cap), and then synthesizing a single answer from the per-chunk candidates in a reduce step. We evaluate on the LongDocURL questions whose document has $100$ pages or more ($817$ questions over $115$ documents, mean $121$ pages) and score with the same official scorer. The comparison point is plain all-images on the identical subset, which scores $0.204$ (against $0.444$ on the $<\!100$-page subset).

Chunked map-reduce reaches $0.193$, against $0.204$ for the single-request baseline as scored elsewhere in the paper (\S\ref{sec:appendix-page-count}), so the two are a wash and chunking neither helps nor hurts (Table~\ref{tab:mapreduce}). Chunking therefore removes the API stitch/downsample without restoring accuracy, and map-reduce remains far below the $0.444$ that all-images achieves on $<\!100$-page documents, so most of the long-document degradation is intrinsic to whole-document image access at this length rather than an artifact of the single request: splitting the document loses the cross-chunk context needed to answer many questions, and the reduce step cannot recover evidence no chunk surfaced. The selective modes (retrieval and the agent), which send only a handful of targeted pages per call regardless of document length, sidestep both failure modes and remain the recommended options on long documents.

\begin{table}[t]
\centering
\scriptsize
\setlength{\tabcolsep}{4pt}
\begin{tabular}{@{}lcc@{}}
\toprule
Configuration & AVG & $\Delta$ vs.\ map-reduce \\
\midrule
Chunked map-reduce                                & 0.193 & --- \\
Single-request all-images, as scored in the paper & 0.204 & $-1.1$\,pp \\
\bottomrule
\end{tabular}
\caption{Chunked map-reduce against single-request all-images on the LongDocURL questions whose documents have $100$ pages or more ($817$ questions over $115$ documents, mean $121$ pages), Sonnet 4.5 under the official scorer. Chunking is a wash, and both configurations sit far below the $0.444$ that all-images reaches on documents under $100$ pages, so the long-document collapse is mostly intrinsic to whole-document image access rather than an artifact of the per-request cap.}
\label{tab:mapreduce}
\end{table}

\end{document}